\documentclass[acmsmall]{acmart}
\renewcommand\footnotetextcopyrightpermission[1]{}
\setcopyright{none}

\usepackage{tikz}
\usetikzlibrary{arrows.meta,positioning,fit,shapes.geometric,calc}
\usepackage{amsmath,mathtools}
\usepackage{amsthm}
\usepackage{graphicx}
\graphicspath{{./}{figures/}}
\usepackage{xcolor}
\usepackage{subfigure}
\usepackage{booktabs}
\usepackage{adjustbox}
\usepackage{algorithm}
\usepackage{algorithmic}

\usepackage[capitalize,noabbrev]{cleveref}

\let\aa\relax

\newcommand{\va}{\mbox{${\mathbf a}$}}

\newcommand{\vu}{\mbox{${\mathbf u}$}}

\newcommand{\mR}{\mbox{${\mathbf R}$}}

\newcommand{\Rh}{\mbox{${\hat R}$}}

\newcommand{\ga}{\alpha}

\newcommand{\gd}{\delta}

\newcommand{\gp}{\pi}

\newcommand{\gr}{\rho}

\newcommand{\gt}{\tau}

\newcommand{\gD}{\Delta}

\newcommand{\bbP}{{\mathbb{P}}}

\newcommand{\bea}{\begin{array}}
\newcommand{\ena}{\end{array}}
\newcommand{\bds}{\begin {itemize}}
\newcommand{\eds}{\end {itemize}}
\newcommand{\bdf}{\begin{definition}}
\newcommand{\blm}{\begin{lemma}}
\newcommand{\edf}{\end{definition}}
\newcommand{\elm}{\end{lemma}}
\newcommand{\bthm}{\begin{theorem}}
\newcommand{\ethm}{\end{theorem}}
\newcommand{\bprp}{\begin{prop}}
\newcommand{\eprp}{\end{prop}}
\newcommand{\bcl}{\begin{claim}}
\newcommand{\ecl}{\end{claim}}
\newcommand{\bcr}{\begin{coro}}
\newcommand{\ecr}{\end{coro}}
\newcommand{\bquest}{\begin{question}}
\newcommand{\equest}{\end{question}}

\newcommand{\nin}{{\not \in}}

\providecommand{\Reg}{\operatorname{Reg}}

\theoremstyle{plain}
\newtheorem{theorem}{Theorem}[section]

\newtheorem{lemma}[theorem]{Lemma}

\theoremstyle{definition}
\newtheorem{definition}[theorem]{Definition}

\theoremstyle{remark}

\begin{document}
\title[Near-Optimal Privacy-Preserving Learning for Max-Min Fair Multi-Agent Bandits]{Near-Optimal Privacy-Preserving Learning for Max--Min Fair Multi-Agent Bandits}
\author{Amir Leshem}
\authornote{This research is partially funded by ISF grant 2197/22.}
\affiliation{
  \institution{Faculty of Engineering, Bar-Ilan University}
  \city{Ramat-Gan}
  \country{Israel}
}
\email{amir.leshem@biu.ac.il}

\ccsdesc[500]{Theory of computation~Self-organization}
\ccsdesc[500]{Theory of computation~Multi-agent learning}
\ccsdesc[500]{Theory of computation~Sequential decision making}
\ccsdesc[500]{Theory of computation~Regret bounds}
\ccsdesc[500]{Theory of computation~Computational pricing and auctions}
\begin{abstract} 
We study fair multi-agent multi-armed bandit learning under collision-only coordination. Agents cannot communicate explicitly during learning and observe only their own rewards and whether collisions occur when several agents access the same arm. The goal is to learn a max-min fair allocation while keeping each agent's reward samples and empirical reward estimates local. We propose a fully distributed algorithm for bounded rewards with unknown support, achieving regret $O\!\left(N^3 f(\log T)\log T\right)$, where $f$ is any nondecreasing diverging function satisfying $f(k-1)/f(k)\to 1$. The algorithm combines distributed agent ordering, cumulative round-robin exploration, endpoint-revalidated warm-started bisection, and a collision-based distributed auction for threshold-feasibility tests. Unlike leader-based optimal algorithms, no agent collects the reward observations, empirical estimates, or preferences of the others. Thus, the protocol preserves reward privacy in the operational sense of avoiding reward sharing, while coordinating only through collision outcomes. Compared with previous privacy-preserving algorithms for max--min fair bandits, which have exponential dependence on the number of agents, our method achieves polynomial $N^3$ dependence while retaining near-logarithmic dependence on $T$. The analysis uses concentration of cumulative empirical estimates and stability of endpoint-revalidated bisection. Simulations confirm the predicted scaling with horizon, number of agents, and max--min gap across representative numerical settings.
\end{abstract}
\maketitle

\section{Introduction}
Large-scale systems often need to allocate indivisible or congestible resources
among multiple agents with heterogeneous and initially unknown utilities.
Examples include assigning jobs to cloud servers, channels to wireless users,
vehicles or routes in transportation systems, and service opportunities in
shared platforms. A classical approach is centralized allocation: a planner
collects the agents' utility information and computes an allocation. However,
as the number of agents and resources grows, collecting and maintaining this
information becomes costly, scaling with the product of the number of agents and
resources. Moreover, agents may not know their own utilities in advance and may
need to learn them through repeated interaction with the system.

This paper studies fair allocation of indivisible resources when agents’ valuations are initially unknown and can only be learned through repeated use. Unlike classical fair division, utilities are not given as input; unlike standard mechanism-design models, the agents are cooperative but privacy-constrained. The contribution is an online-learning analogue of max–min fair allocation with a decentralized implementation that does not require agents to reveal reward samples, empirical utility estimates, or preferences.

These considerations motivate distributed learning protocols in which agents
make local decisions, learn their own utilities, and coordinate through minimal
public feedback. Such protocols are especially relevant when utilities are
private or costly to communicate. The challenge is to design a learning rule that
allows the agents to converge to a fair allocation without requiring a central
planner to collect reward samples, empirical utility estimates, or preference
information from the agents.
The multi-agent multi-armed bandit setup is a good paradigm for distributedly allocating resources under uncertainty \cite{bistritz2018distributed}. 
This emerging paradigm for multi-agent resource allocation has been recently the subject of extensive research efforts, see, e.g. \cite{liu2010distributed,xu2015distributed},
\cite{Vakili2013}, \cite{Lai2008}, \cite{Anandkumar2011}, \cite{liu2019competing}, \cite{Liu2013}, \cite{Avner2014}, \cite{nayyar2016regret}, \cite{Evirgen2017}, \cite{Cohen2017}, \cite{Avner2016}, \cite{zafaruddin2019distributed}, \cite{hanawal2018multi}, \cite{ bistritz2020cooperative}.
 Early works concentrated on the special case where each arm has the same reward distribution irrespective of the agent selecting it and optimized the sum of rewards \cite{Rosenski2016}, \cite{boursier2019sic}, \cite{alatur2020multi}, \cite{bubeck2019non}. However, in many applications, the reward each arm provides to different agents might be different, e.g. in the context of cloud computing, certain users or applications might benefit from using machines with faster CPU, while others will benefit more from having access to machines with faster and larger memory. Similarly, when allocating wireless channels, different receivers might experience different interference at different frequencies, resulting in different achievable data rates. 

In the standard model, each agent faces the classical stochastic multi-armed bandit problem \cite{bubeck2012regret}, but the agent is impacted by the choices of all players. 
The assumed model is fully cooperative, where agents are allowed to set up a joint protocol in advance, but not explicitly send messages to each other during the learning phase. Although the protocol is cooperative, agents are unaware of the other agents' actions and rewards, which can result in conflicting actions. 

One approach to resolve this problem is by assigning zero rewards to players who select the same arm, or equivalently, providing the agents with a collision indicator, similar to the collision/failed-ACK message in wireless networks. Furthermore, in terms of network protocols, this is equivalent to implementing a multichannel ALOHA protocol, where each arm (resource) has a dedicated channel, with shared access. Therefore, by learning which arms to pull, the agents can jointly learn an optimal allocation distributedly. The only information an agent receives is through the collisions occurring when other agents select the same arm. This collision model captures Aloha-based protocols in communication networks, computation resources on servers, consumers splitting indivisible goods, etc. An alternative to this approach, which alleviates the need for message passing, is by applying a distributed auction algorithm \cite{naparstek2013fully} together with an access protocol based on opportunistic carrier sensing \cite{zhao2005opportunistic}. This generalization has been proposed by \cite{zafaruddin2019distributed} and later extended to learning how to share the same arm distributedly among multiple players when incorporating more agents than arms \cite{boyarski2023distributed}. 

Initial works on the problem focused on maximizing the total sum of agent rewards, e.g.,  \cite{hanawal2018multi}, \cite{Besson2018}, \cite{tibrewal2019distributed}, \cite{bistritz2018distributed}, \cite{bistritz2021game}, \cite{Kalathil2014}, \cite{nayyar2016regret}, \cite{boursier2019sic}, \cite{boursier2019practical}.
 The total sum utility metric is relevant in some applications when only the overall system utility is considered.
 However, in distributed environments where agents have conflicting interests while they still benefit from cooperation, there is a need to motivate all agents to collaborate and achieve an egalitarian allocation. For example, any division should be individually reasonable, so that each agent receives more than what it can gain by competing.  In the general cooperative game-theoretical literature, fair solutions are important means to facilitate cooperation. Examples are the seminal Nash Bargaining Solution \cite{nash1950bargaining}, proportional fair division \cite{kubiak2008proportional}, Kalai-Smorodinski solution \cite{kalai1975other} and weighted max-min fair allocations \cite{mjelde1983properties}, \cite{zehavi2013weighted}.  These objectives have been used extensively in the broader resource allocation literature   \cite{radunovic2007unified,zehavi2013weighted,asadpour2010approximation}. In contrast, fairness in multi-player bandits has only recently been studied \cite{bistritz2020my}, \cite{bistritz2021one}. Some multi-player bandit works have studied alternative objectives
that can potentially exhibit some level of fairness, e.g., \cite{darak2019multi,bar2019individual}. Several recent works have studied fairness-constrained sequential learning for a single player \cite{jabbari2017fairness,joseph2016fairness,zhang2019group}.
The work by Bistritz et al. \cite{bistritz2021one} also considered a quality of service guarantee, when it is known that the expected level of service is achievable. Surprisingly, in this setting, regret is bounded. However, assuming knowledge of the feasibility of a given QoS level is strong, and not always practical. These works provide regret bounds and achievable regret for the fair multi-armed bandit problem. 
Unfortunately, from the computational point of view, they involve a very large state space Markov chain of size $(2K)^N$, where $N$ is the number of players, and $K$ is the number of arms. Even for $2$ arms, the state space is of size $4^N$. 

In this paper, we consider max-min fair solutions to the multiplayer
multi-armed bandit problem. The goal is to learn an assignment that maximizes
the expected reward of the worst-off agent. We propose a fully distributed
algorithm that uses only collision-based coordination and does not require any
agent to share reward samples, empirical reward estimates, or preferences. For
bounded rewards with unknown support, the algorithm achieves regret
\[
    O\!\left(N^3 f(\log T)\log T\right),
\]
where \(f(t)\) is any nondecreasing function diverging to infinity satisfying $\frac{f(k-1)}{f(k)} \to 1$. The
dependence on the horizon is therefore near-logarithmic, while the dependence on
the number of agents is polynomial. This improves over previous fair
multi-player bandit algorithms that achieve near-logarithmic regret in \(T\) but
have exponential dependence on the number of agents through a large Markov-chain
state space. The \(T\)-dependence is essentially tight, since even the
single-player stochastic multi-armed bandit problem has an \(\Omega(\log T)\)
regret lower bound \cite{lai1985asymptotically}. 
This improves prior art, which provided a near-optimal regret order but also had exponential dependence on the number of agents. Moreover, the polynomial dependence makes the proposed technique attractive in large-scale problems.

Interestingly, a simple variation of our learning algorithm provides a learning algorithm achieving any Pareto dominant allocation, through the use of weighting, similarly to the results of \cite{zehavi2013weighted}.

\subsection{Prior Work on fair bandit learning}
As mentioned, problems of fairness date back to Nash's work on the bargaining problem \cite{nash1950bargaining}. A special case of the Nash bargaining solution is the proportional fair solution, which is equivalent to Nash's solution when the disagreement value is $0$ for all players and leads to maximizing $\sum_n \log U_n(\va)$ where $\va$ is the vector of the joint actions. Another family of fairness criteria is defined by the 
\textit{$\alpha$-fairness} metric, for a vector of rewards $\vu \in {\mathbb{R}}^N$ is $\sum_{i=1}^N (1-\alpha)^{-1} u_i^{1-\alpha}$, for $\ga\neq 1$ \cite{mo2000fair_alpha}.
This notion of fairness encompasses several classical ones, where $\alpha=0$ yields the sum of rewards, and the max-min fairness criterion corresponds to the limit as $\alpha\rightarrow \infty$.
For a constant $\alpha$, $\alpha$-fairness can be maximized in a similar manner to \cite{bistritz2018distributed,bistritz2021game}, the case of max-min fairness is fundamentally different.

When analyzing max-min fair learning regret, it is simple to show an $\Omega(\log T)$ lower bound using a reduction to the single player Lai-Robbins lower bound by adding fictitious players with high rewards (See \cite{bistritz2021one}, proposition 1), so the main problem is to find tight upper bounds on the problem.

Learning to play a max-min fair allocation without explicit communication between the players poses several challenges that do not arise in the case of maximizing the sum-rewards (or in the case of \textit{$\alpha$-fairness}). The sum-rewards
optimal allocation is unique for ``almost all'' scenarios (by dithering the expected rewards). In contrast, there are typically multiple max-min fair allocations.
 This complicates the distributed learning process since players will have to agree on a specific optimal allocation to play, which is difficult to do without communication. Specifically, this rules out using similar techniques to those used in \cite{bistritz2018distributed} to solve the sum of rewards case. The first paper to propose a solution was \cite{bistritz2020my} and its extension \cite{bistritz2021one}. The first result in these papers proved near logarithmic regret for the max-min problem. Interestingly, in the second paper, it was also proved that when the value of the max-min is known, a bounded regret can be achieved by extending the QoS formulation in \cite{lai1984asymptotically} and \cite{katz2020true}, which found all the ``good arms'' (instead of minimizing the regret). However, neither \cite{lai1984asymptotically} nor \cite{katz2020true} can be used for the multiplayer case since they rely on i.i.d. rewards, which is no longer the case with collisions between players.  In \cite{lai1984asymptotically}, it is proved that if a number $\gamma$ between the optimal expected reward and the second-best expected reward is known, then $O(1)$ regret can be achieved for the single-player multi-armed bandit problem. This was extended to the multiplayer case in \cite{bistritz2021one}. As explained above, the techniques of \cite{bistritz2021one} become exponentially complex as the number of agents grows, because it relies on the convergence of an absorbing Markov chain with an exponentially large (in the number of players) state space. 

In contrast, our approach separates the statistical learning problem from the
distributed computation of the fair allocation. We first use a preliminary
distributed ordering phase, a standard task in distributed computation, to assign
agents distinct roles. This ordering prevents unnecessary random collisions in
later phases and allows collisions to be used as controlled signals. Once the
agents are ordered, each epoch consists of exploration, distributed matching, and
exploitation. During the matching phase, the agents use their current empirical
reward estimates to test whether a candidate max-min threshold is feasible. Each
threshold test is implemented by a collision-based distributed auction procedure,
so no rewards or empirical reward estimates are exchanged.

This use of distributed auctions is related to prior work on decentralized
multi-player bandits. The auction algorithm was used in~\cite{nayyar2016regret}
to obtain logarithmic regret when communication between agents is possible, and
a fully distributed auction-based method was later proposed in
\cite{zafaruddin2019distributed} for sum-rate maximization. However, that work
relies on listen-before-talk access, which is analogous to replacing ALOHA by
CSMA in wireless networks~\cite{rom2012multiple}, and it assumes discrete rewards.
Here we work in the collision-only model and target the max-min fairness
objective. The resulting algorithm replaces the exponentially large Markov-chain
matching dynamics used in earlier fair-bandit algorithms by a polynomial-time
distributed feasibility test, leading to near-logarithmic regret in the horizon
with polynomial dependence on the number of agents. 

\subsection{Contributions and limitations}

The main contribution of this paper is a fully distributed algorithm for learning
a max-min fair allocation in a heterogeneous multi-agent multi-armed bandit
problem under collision-only coordination. Agents agree on the protocol in
advance, but during learning, they do not exchange reward samples, empirical
reward estimates, prices, or preference information. Coordination is achieved
only through collisions and through the common feasible/infeasible outcomes of
distributed threshold tests.

The algorithm combines ideas from online learning and distributed computation.
First, a short distributed ordering phase assigns agents distinct roles. This
prevents unnecessary random collisions in later phases and enables collisions to
be used as controlled signals. Second, during each epoch, agents perform
fixed-length round-robin exploration, so that each agent forms cumulative
empirical estimates of its own rewards for all arms. Third, the agents compute a
max-min allocation for the current empirical reward matrix by running a
distributed bisection over threshold values. For a threshold \(\tau\), feasibility
means that the bipartite graph containing edges
\[
    (n,m) \quad \text{such that} \quad \widehat R_{n,m}(k)\ge \tau
\]
contains a perfect matching. Each feasibility test is implemented by a
collision-based distributed auction procedure.

The distributed auction component builds on the auction algorithm of
Bertsekas~\cite{bertsekas1979distributed} and on the fully distributed variant
of~\cite{naparstek2013fully}, in which each agent maintains its own local prices.
For the maximum-cardinality matching problem considered here, the auction can be
viewed as a push-relabel or double-push procedure
\cite{goldberg1995efficient,bertsekas1992forward,naparstek2016expected}. The
initial ordering step lets the agents schedule auction operations without a
central coordinator. This gives a distributed feasibility test with polynomial
complexity, of order \(O(N^3)\) slots for an \(N\times N\) threshold graph.

A key technical point is that the empirical reward matrix changes from epoch to
epoch, so a threshold bracket that was valid in one epoch need not remain valid
in the next. Distributed bisection is therefore delicate: the agents must
maintain a common threshold bracket without exchanging numerical information.
We address this through endpoint-revalidated warm-started bisection. At the
beginning of each matching phase, the agents retest the previous lower and
upper endpoints using the current empirical matrix. If the bracket is no longer
valid, it is repaired using only distributed feasibility outcomes. The agents
then continue bisection from the repaired bracket. This mechanism keeps the
threshold search synchronized without requiring communication of reward
estimates or bracket corrections.

In contrast to approaches that use coded collisions to communicate reward
samples or empirical reward estimates, our protocol uses collisions only as
feasibility signals. No agent reconstructs or knows another agent's observed
rewards or empirical utility estimates; these quantities remain local to the
agent throughout learning. Thus, the algorithm preserves reward privacy in the
operational sense that reward information is never explicitly shared, while
still enabling the agents to agree on a common fair allocation.

The resulting regret bound is
\[
    O\!\left(N^3 f(\log T)\log T\right),
\]
where $f$ is any nondecreasing function satisfying
$f(k)\to\infty$ and $f(k-1)/f(k)\to 1$. Hence the
algorithm preserves the near-logarithmic dependence on the horizon obtained in
prior fair-bandit algorithms, while replacing their exponential dependence on the number of agents by polynomial dependence. The proof relies on the concentration
of cumulative empirical reward estimates, on the gap between the optimal and
best non-optimal max-min assignments, and on the stability of the
endpoint-revalidated bisection procedure.

The main limitation is that the regret is near-logarithmic rather than exactly
logarithmic. As in~\cite{bistritz2021one}, the bound has the form
\(f(\log T)\log T\), where \(f(t)\) may be chosen to diverge arbitrarily slowly.
We also assume bounded rewards. The agents do not need to know the reward bound,
but boundedness is used in the concentration analysis. Extending the result to
general sub-Gaussian rewards while retaining a simple polynomial bound on the
distributed matching phase remains open.
\section{The max-min fair bandit problem}
 Assume that $N$ agents access $M$ arms with agent-dependent mean rewards $R_{n,m}$. The agents do not know the mean rewards and cannot communicate with each other. They need to learn the optimal arm assignment. Time $t$ is discrete and synchronized. Each time an agent $n$ chooses an arm $m$ she obtains a random reward $r_{n,m}$. When two agents access the same arm simultaneously, a collision occurs, and the reward of the colliding agents is $0$. We define for each arm and action profile $\va=[a_1,\ldots,a_N]$, where $a_n$ is the arm selected by agent $n$, a collision indicator $\eta_{m}(\va)$  by:
 \begin{align}
     \eta_{m}(\va)=\left\{
     \begin{tabular}{ll}
         $1$ & \hbox{if \ } $\left|\left\{n:a_n=m\right\}\right|= 1$.\\
         $0$ & \hbox{otherwise.}
     \end{tabular}
     \right.
\end{align}
 The instantaneous utility of agent $n$  at time $t$ is now given by 
 $u_n(t)=\eta_{a(t)}(t)r_{n,a_n}(t)$.
The agents are cooperating in the sense that they can follow a predefined shared protocol, but they cannot exchange information regarding the rewards they collected or their preferences. For simplicity of exposition, we assume that $N=M$, since the availability of extra arms simplifies the coordination process. In the concluding remarks of the paper, we discuss the cases $M>N$ and $M<N$. 
The max-min allocation problem can be described as finding the value $\gr^*$ and permutation $\pi^*$ which satisfy: 
\begin{align}
      \gr^*&=\max_{\gp\in S_n} \min_n R_{n,\gp(n)}\\
      \pi^*&=\arg \max_{\gp\in S_n} \min_n R_{n,\gp(n)}
\end{align}
We now define the set of optimal allocations, since there might be multiple optimal allocations. 
Let the set of optimal allocations be
\begin{align}
  A_1=\left\{\pi: \min_{n} R_{n,\pi(n)} =\gr^* \right\}.
\end{align}
Without loss of generality, we assume $A_1\neq S_n$, since then the problem trivializes.
Define the gap $\gD$ by
\begin{align}
    \gD=\gr^*-\gr_2
\end{align}
where $\gr_2=\max_{\gp\nin A_1} \min_n R_{n,\gp(n)}$
We can now define the (pseudo)-regret of a strategy $\pi$ satisfying $\pi(t)=(a_1(t),\ldots,a_N(t))$ as:
\begin{align}
    \Reg(T)
    =
    \sum_{t=1}^T
    \left[
        \rho^*
        -
        \min_{n\in[N]}
        \eta_{a_n(t)}(\mathbf a(t))R_{n,a_n(t)}
    \right].
\end{align}

 In this paper, we assume that the sample rewards are positive and bounded random variables, but their support is unknown to the agents. For a distribution $F(r)$, 
 we denote by $B(F)$ the upper endpoint of the support,
\[
B(F)=\inf\{x:F(x)=1\}.
\]
When $F$ is the distribution of the reward of agent $n$ on arm $m$ we denote this by $B_{n,m}$.
We do not assume identical support of the arm rewards: For each agent $n$ and each arm, we can have different support $B_{n,m}$. 

{\em{Bounded rewards}:}
While theoretically, we would be interested in the general sub-Gaussian case, from a practical point of view this limiting assumption always holds for physical reasons. Hence, the bounded rewards with unknown support are a reasonable model. We will also assume that the distribution of the arm rewards is continuous with positive density on the support of the distribution. This assumption is not necessary, but it simplifies the notation and, therefore, the presentation. The techniques in this paper do not carry straightforwardly to the general sub-Gaussian case, leaving this case as an interesting research problem. 

{\em Continuously distributed rewards:}
Rewards with a continuous distribution are natural in many applications (e.g., signal-to-noise ratio in wireless networks). However,
this assumption is only used to argue that since the probability of
zero reward in a non-collision is zero, players can properly estimate their expected rewards. In the case
where there is a positive probability of receiving zero reward, we can assume
instead that each player can observe their no-collision indicator in addition
to their reward. This alternative assumption requires no modifications to our algorithms or analyses. 
Observing one bit of feedback signifying whether any other player chose
the same arm is significantly less than other common feedback models, such as observing the actions of other players. In wireless networks, this could mean that the \texttt{ACK} is not received at the transmitter over the reverse control channel and therefore the transmitter knows there was a collision on its chosen channel.

 \section{Learning a max-min  optimal allocation}
Similarly to other cooperative learning schemes that operate under limited communication, we will use phased exploration, 
negotiation and learning. 
In contrast to previous work, we would like to order the agents by adding an agent-ordering phase at the beginning of the learning process.
This makes the exploration more efficient and allows using collisions to determine the end of phases. This replaces the leader election process in algorithms where the leader collects the rewards. However, the implementation has a fixed complexity, slightly larger than the leader election protocol.

The cumulative exploration phase length will grow faster than linearly and is $LKf(k)$ at epoch $k$.

In the negotiation phases, an optimal matching based on the current estimates of the rewards is computed in a distributed manner. To achieve this, we propose a novel collision-only matching protocol for
the fair bandit setting. The protocol uses the distributed auction/push-relabel
matching procedure as a primitive, but deploys it as a threshold-feasibility
oracle inside an endpoint-revalidated max-min bisection scheme. This
combination allows the agents to compute a fair allocation without sharing
reward samples, empirical means, prices, or preference information. Finally, the length of the exploitation phase will grow exponentially.
 
\begin{figure}[t]
    \centering
    \includegraphics[width=\textwidth]{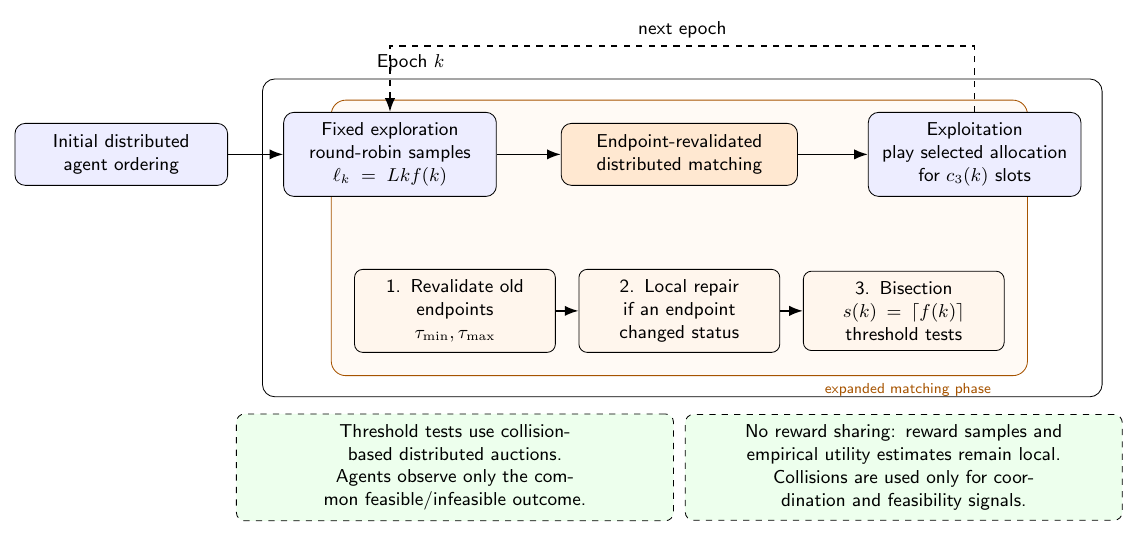}
    \caption{Overview of the proposed collision-only fair bandit algorithm.
    After an initial distributed ordering phase, the algorithm proceeds in epochs.
    Each epoch consists of fixed-length round-robin exploration, an endpoint-revalidated
    distributed matching phase, and exploitation of the last feasible allocation.
    The matching phase maintains a common threshold bracket by revalidating the
    previous endpoints, repairing the bracket if needed, and then running bisection.
    Each threshold feasibility test is implemented by a collision-based distributed
    auction. Thus, agents coordinate only through collision outcomes and never
    exchange reward samples, empirical utility estimates, or numerical messages.}
    \label{fig:algorithm_overview}
\end{figure}

Since the support of each arm is different, we will need a coordination mechanism to end the exploration and exploitation phases. A similar approach will be used to determine distributedly that the max-min allocation using the estimated rewards of each agent is achieved.
The algorithm is given in \ref{alg:FairBandits} and in Figure~\ref{fig:algorithm_overview}.
\subsection{Agent's ordering}
As a first step, we would like to determine an agreed order of the agents in order to simplify the analysis of the protocol.
Therefore, we assume that arms are ordered according to some random order, known to the players. This is not a limiting assumption, since the arms are entities that can have a fixed identifier, which allows the agents to select specific arms. Using this assumption, we can use the collision mechanism on the arms to order the agents randomly. Moreover, each agent only knows her own rank. We can replace the pre-ordering phase with random access, but this will complicate the analysis of the subsequent phases, without any significant gain. As we will show below, this step has a low constant regret of the order $O\left(N\log N \right)$.
Ordering agents is a standard symmetry-breaking task in distributed
computation, closely related to randomized leader election
\cite{Ramanathan2007RandomizedLeaderElection}. In our collision model,
we use the same symmetry-breaking principle to assign agents distinct
temporary ranks rather than to elect a single coordinator.
The ordering phase consists of repeated ordering blocks of length
\(O(N\log N)\), followed by a collision-based termination check. A standard coupon-collector argument shows that the bottleneck case, in which only one
agent remains unassigned, terminates within \(4N\log_2 N\) trials with
probability at least \(1-N^{-4/\ln 2}\) (See Appendix~\ref{app:agent_ordering}).

Following the initial ordering phase, the algorithm is divided into epochs of varying lengths. 
As described in Algorithm~\ref{alg:FairBandits} the learning continues in epochs, each comprised of 3 phases, similarly to \cite{bistritz2021game}.
\begin{algorithm}[]
\begin{algorithmic}[1]
\caption{\label{alg:FairBandits} \texttt{The optimal fair bandit algorithm}}
\STATE \textbf{Initialization of agent $n$}: 
\STATE $N$ is the number of agents.
\STATE $M$ is the number of arms.
\STATE $\Rh_{n,m} \gets 0$ for all $m$.
\STATE $V_{n,m}\gets 0$ for all $m$.
\STATE Each agent performs Agent ordering$(n)$:
\FOR{epoch $k=1,2,\ldots$} 
\STATE Each agent $n$ performs Exploration$(k,n)$.
\STATE Each agent $n$ performs Matching $(k,n)$.
\STATE Each agent $n$ performs Exploitation$(k,n)$.
\ENDFOR
\end{algorithmic}
\end{algorithm}
Below, we provide the details of the fair distributed learning algorithm. Detailed pseudo code is provided in Appendix~\ref{app:code}.
\subsection{Exploration Phase}
\label{sec:Exploration}
We begin with the exploration phase. In this phase, the players sample the arms to obtain an unbiased estimate of the rewards. They also use estimates from previous epochs. The length of the exploration phase is sufficiently large to ensure that the estimates are sufficiently accurate so that the probability of error in the matching phase is sufficiently low to bound the regret during the exploitation phase. 
Since we have shown that the agents can be ordered with bounded regret and each agent is assigned to an arm, without loss of generality, we can assume that there are no collisions during the exploration phase, since each agent applies a round-robin schedule, initialized at its uniquely selected arm.  Over time, agents receive stochastic rewards from different arms and average them to estimate their expected reward for each arm. During each
epoch, $k$,  $c_{1}(k)=L\left(g(k)-g(k-1)\right)$ slots are dedicated to the exploration of each arm, where $f(k)$ is any nondecreasing function satisfying $f(k)\to\infty$ and $f(k-1)/f(k)\to 1$ and $g(k)=kf(k), g(0)=0$ and  $L$ is a meta-parameter used to improve the convergence rate. 

More generally, the analysis applies to any nondecreasing function
$f(k)\to\infty$.
Note that by the choice of $c_1(k)$,  and the round-robin scheduling of the exploration, the total number of exploration steps of each arm until the end of the $K$'th exploration phase is 
\begin{align}
\label{def:lk}
    \small \ell_K=\sum_{k=1}^K c_1(k)/M=LKf(K),
\end{align}
since this is a telescopic series. 

The purpose of the exploration phase is to help the players become more confident over time regarding the value of the arms, so that finding a max-min assignment with respect to the estimated arms will eventually correspond to the optimal assignment with respect to the true estimates. We note that the distributed ordering of the agents improves the performance compared to the exploration in
\cite{Rosenski2016,bistritz2018distributed, bistritz2020my}. 
The exploration phase is described in Algorithm~\ref{alg:exploration} in the appendix.


\subsection{Matching phase}
\label{sec:matching_revised}
The matching phase is used to distributedly compute a max-min allocation using the
current empirical reward matrix. We begin with a high-level description and then
explain how each feasibility test is implemented using collisions.

As a first step, we recall the connection between max-min allocations and perfect
matchings in threshold graphs.
\begin{lemma}
\label{lem:bipartite_bound}
   Let $\mR$ be a reward matrix. For every threshold $\gt$, $\gt\le \gr^*$, where
   $\gr^*$ is the optimal max-min value, if and only if the bipartite graph
   $G_R(N,N,\gt)=\left(V_1,V_2,E(\gt)\right)$ defined by
\begin{align}
        V_1=\{1,\ldots,N\},  \quad  V_2=\{1,\ldots,N\}, \\
       E(\gt) = \left\{(n,m): n \in V_1, m\in V_2,  \gt \le R_{n,m}\right\}
   \end{align}
   has a perfect matching.
\end{lemma}
The proof is immediate. If $\gt\le \gr^*$, any max-min optimal permutation gives a
perfect matching in $G_R(N,N,\gt)$. Conversely, a perfect matching in
$G_R(N,N,\gt)$ with $\gt>\gr^*$ would contradict the definition of $\gr^*$.
When $N<M$, the same statement holds with a one-to-one matching from agents to
arms.

Based on Lemma~\ref{lem:bipartite_bound}, the agents search over thresholds
$\tau$ and test whether the empirical graph $G_{\Rh(k)}(N,N,\tau)$ admits a
perfect matching. Each feasibility test is implemented by the distributed auction
protocol of Section~\ref{subsec:dist_auction}.

\paragraph{Warm-started bisection with endpoint revalidation.}
The bisection routine is warm-started across epochs. Since all agents observe the
same feasibility or infeasibility outcome of each distributed matching test, the
stored bracket
\[
    [\tau_{\min}(k-1),\tau_{\max}(k-1)]
\]
is common to all agents. However, the empirical reward matrix changes from
$\Rh(k-1)$ to $\Rh(k)$, and therefore the old bracket is not assumed to remain
valid automatically.

At the beginning of the matching phase in epoch $k$, the agents first run two
distributed feasibility tests using the current empirical matrix $\Rh(k)$: one at
$\tau_{\min}(k-1)$ and one at $\tau_{\max}(k-1)$.

There are three cases.

First, if $\tau_{\min}(k-1)$ is feasible and $\tau_{\max}(k-1)$ is infeasible,
then the old bracket remains valid for the current empirical matrix, and the algorithm continues bisection from
\[
    [\tau_{\min}(k-1),\tau_{\max}(k-1)].
\]

Second, if $\tau_{\max}(k-1)$ is feasible, then the current empirical max-min
value lies above the old bracket. Let
\[
    \eta_{k-1}\triangleq \tau_{\max}(k-1)-\tau_{\min}(k-1).
\]
The agents set
\[
    \tau_{\min}\leftarrow \tau_{\max}(k-1),
\]
and test the thresholds
\[
    \tau_{\max}(k-1)+\eta_{k-1},
    \quad
    \tau_{\max}(k-1)+2\eta_{k-1},
    \quad
    \tau_{\max}(k-1)+4\eta_{k-1},
    \ldots
\]
until an infeasible threshold is found. This produces a valid bracket for
$\Rh(k)$.

Third, if $\tau_{\min}(k-1)$ is infeasible, then the current empirical max-min value lies below the old bracket. The agents set
\[
    \tau_{\max}\leftarrow \tau_{\min}(k-1),
\]
and test the thresholds
\[
    \max\{0,\tau_{\min}(k-1)-\eta_{k-1}\},
    \quad
    \max\{0,\tau_{\min}(k-1)-2\eta_{k-1}\},
    \quad
    \max\{0,\tau_{\min}(k-1)-4\eta_{k-1}\},
    \ldots
\]
until a feasible threshold is found. Since rewards are nonnegative, a threshold
zero is feasible, so this repair step always terminates.

After endpoint revalidation and, if needed, local repair, the agents perform
\[
    s(k)\triangleq \lceil f(k)\rceil
\]
additional bisection feasibility tests.

To avoid storing brackets that are unnecessarily narrower than the next-epoch
empirical drift, the stored bracket is padded to a public minimum width
\[
    w_k\triangleq 2^{-f(k)}.
\]
That is, if the final bracket after bisection has width smaller than $w_k$, the
agents decrease the lower endpoint and increase the upper endpoint so that the stored bracket has a width of $w_k$. This preserves the validity of the bracket because
decreasing the lower endpoint preserves feasibility and increasing the upper
endpoint preserves infeasibility.

The procedure uses only distributed auction outcomes. No rewards, empirical means, or numerical messages are exchanged between agents.

\begin{lemma}[Stability of the empirical max-min value]
\label{lem:empirical_value_stability}
Let
\[
    \hat\rho_k^*
    \triangleq
    \max_{\pi\in S_N}\min_n \Rh_{n,\pi(n)}(k)
\]
be the empirical max-min value at epoch $k$. Define
\[
    D_k
    \triangleq
    \max_{n,m}
    \left|
        \Rh_{n,m}(k)-\Rh_{n,m}(k-1)
    \right|.
\]
Then
\[
    \left|\hat\rho_k^*-\hat\rho_{k-1}^*\right|
    \le D_k.
\]
Moreover, since the estimates are cumulative,
\[
    D_k
    \le
    B\frac{\ell_k-\ell_{k-1}}{\ell_k},
    \qquad
    \ell_k=Lk f(k).
\]
\end{lemma}

\begin{proof}
For any fixed assignment $\pi$,
\[
    \left|
        \min_n \Rh_{n,\pi(n)}(k)
        -
        \min_n \Rh_{n,\pi(n)}(k-1)
    \right|
    \le D_k.
\]
Taking the maximum over $\pi\in S_N$ preserves the same Lipschitz bound, hence
\[
    \left|\hat\rho_k^*-\hat\rho_{k-1}^*\right|
    \le D_k.
\]

For the second claim, the estimates are cumulative:
\[
    \Rh_{n,m}(k)
    =
    \frac{\ell_{k-1}}{\ell_k}\Rh_{n,m}(k-1)
    +
    \frac{\ell_k-\ell_{k-1}}{\ell_k}
    \bar r^{\,\mathrm{new}}_{n,m}(k),
\]
where $\bar r^{\,\mathrm{new}}_{n,m}(k)\in[0,B]$. Therefore
\[
    \left|
        \Rh_{n,m}(k)-\Rh_{n,m}(k-1)
    \right|
    \le
    B\frac{\ell_k-\ell_{k-1}}{\ell_k}.
\]
\end{proof}

\begin{lemma}[Endpoint revalidation and local repair]
\label{lem:endpoint_revalidation_repair}
Suppose that
\[
    [\tau_{\min}(k-1),\tau_{\max}(k-1)]
\]
is a valid bracket for $\Rh(k-1)$, and let
\[
    \eta_{k-1}
    =
    \tau_{\max}(k-1)-\tau_{\min}(k-1).
\]
After the endpoint revalidation and local-repair step described above, the agents
obtain a valid bracket for $\Rh(k)$ whose width is at most
\[
    2(\eta_{k-1}+D_k).
\]
Consequently, after the $s(k)$ bisection tests in epoch $k$, the bracket width
satisfies
\[
    \eta_k
    \le
    \max\left\{
        2^{-f(k)},
        2^{1-s(k)}(\eta_{k-1}+D_k)
    \right\}.
\]
\end{lemma}

\begin{proof}
Let $\hat\rho_k^*$ denote the empirical max-min value at epoch $k$. By
Lemma~\ref{lem:empirical_value_stability},
\[
    \hat\rho_k^*
    \in
    [\hat\rho_{k-1}^*-D_k,\hat\rho_{k-1}^*+D_k].
\]
Since the old bracket was valid,
\[
    \tau_{\min}(k-1)
    \le
    \hat\rho_{k-1}^*
    <
    \tau_{\max}(k-1).
\]

If the old endpoints remain feasible and infeasible, respectively, then no repair
is needed and the starting width is $\eta_{k-1}$.

If $\tau_{\max}(k-1)$ is feasible for $\Rh(k)$, then
$\hat\rho_k^*\ge \tau_{\max}(k-1)$. However,
\[
    \hat\rho_k^*
    <
    \tau_{\max}(k-1)+D_k.
\]
The upward geometric search therefore, finds an infeasible threshold within
distance at most $2\max\{\eta_{k-1},D_k\}$ above $\tau_{\max}(k-1)$.

Similarly, if $\tau_{\min}(k-1)$ is infeasible for $\Rh(k)$, then
$\hat\rho_k^*<\tau_{\min}(k-1)$, while
\[
    \hat\rho_k^*
    \ge
    \tau_{\min}(k-1)-D_k.
\]
The downward geometric search therefore, finds a feasible threshold within
distance at most $2\max\{\eta_{k-1},D_k\}$ below $\tau_{\min}(k-1)$.

Thus, after endpoint revalidation and repair, the bracket width is at most
\[
    2(\eta_{k-1}+D_k).
\]
After $s(k)$ bisection tests, this width is multiplied by at most $2^{-s(k)}$.
Finally, the bracket is padded to width at least $w_k=2^{-f(k)}$, which gives
\[
    \eta_k
    \le
    \max\left\{
        2^{-f(k)},
        2^{1-s(k)}(\eta_{k-1}+D_k)
    \right\}.
\]
\end{proof}
\begin{lemma}[Cost of endpoint revalidation and geometric repair]
\label{lem:geometric_repair_cost}
Assume rewards are supported on \([0,B]\), where \(B<\infty\) is used only in the
analysis. Let
\[
    \ell_k=Lk f(k)
\]
be the cumulative number of samples per agent--arm pair by the end of epoch
\(k\), and suppose that the stored bracket at the end of epoch \(k-1\) has width
at least
\[
    w_{k-1}=2^{-f(k-1)} .
\]
Let \(Q_k\) denote the number of distributed feasibility tests used in epoch $k$ for endpoint revalidation and, if needed, geometric repair, not including
the subsequent \(s(k)=\lceil f(k)\rceil\) bisection tests. Then
\[
    Q_k
    \le
    4+\left\lceil \log_2^+ B\right\rceil+\left\lceil f(k-1)\right\rceil,
    \qquad k\ge2,
\]
where
\[
    \log_2^+ B\triangleq \max\{0,\log_2 B\}.
\]
Consequently, for every horizon \(T\), if
\[
    K_T=\left\lceil \log_2(T+1)\right\rceil ,
\]
then the total number of distributed feasibility tests used for endpoint-revalidation, geometric repair, and bisection over epochs \(2,\ldots,K_T\) is at
most
\[
    C_B K_T + 2K_T f(K_T),
\]
where
\[
    C_B\triangleq 6+\left\lceil\log_2^+ B\right\rceil .
\]
\end{lemma}

\begin{proof}
Endpoint revalidation uses exactly two distributed feasibility tests: one at
\(\tau_{\min}(k-1)\) and one at \(\tau_{\max}(k-1)\).

If the old bracket remains valid, no repair test is needed. Suppose instead that
\(\tau_{\max}(k-1)\) is feasible under the current empirical matrix. Let
\[
    \eta_{k-1}
    =
    \tau_{\max}(k-1)-\tau_{\min}(k-1)
\]
be the stored bracket width. The upward geometric repair tests
\[
    \tau_{\max}(k-1)+\eta_{k-1},
    \quad
    \tau_{\max}(k-1)+2\eta_{k-1},
    \quad
    \tau_{\max}(k-1)+4\eta_{k-1},\ldots
\]
until an infeasible threshold is found.

Let
\[
    D_k
    =
    \max_{n,m}
    \left|
        \widehat R_{n,m}(k)-\widehat R_{n,m}(k-1)
    \right|.
\]
By the stability of the empirical max-min value,
\[
    |\hat\rho_k^*-\hat\rho_{k-1}^*|\le D_k .
\]
Therefore, since the old upper endpoint was above \(\hat\rho_{k-1}^*\), it is
enough to increase the threshold by more than \(D_k\). Hence, the number of
upward repair tests is at most
\[
    2+\left\lceil
        \log_2^+\left(\frac{D_k}{\eta_{k-1}}\right)
    \right\rceil .
\]
The same bound holds for the downward repair when \(\tau_{\min}(k-1)\) becomes
infeasible.

Since the estimates are cumulative and rewards are bounded by \(B\),
\[
    D_k
    \le
    B\frac{\ell_k-\ell_{k-1}}{\ell_k}
    \le B .
\]
Moreover, by the padding rule,
\[
    \eta_{k-1}\ge w_{k-1}=2^{-f(k-1)} .
\]
Thus
\[
    \frac{D_k}{\eta_{k-1}}
    \le
    B2^{f(k-1)} .
\]
Consequently, the number of geometric repair tests is at most
\[
    2+\left\lceil\log_2^+ B\right\rceil+\left\lceil f(k-1)\right\rceil .
\]
Adding the two endpoint revalidation tests gives
\[
    Q_k
    \le
    4+\left\lceil\log_2^+ B\right\rceil+\left\lceil f(k-1)\right\rceil .
\]

Finally, in epoch \(k\), after revalidation and repair, the algorithm performs
\[
    s(k)=\lceil f(k)\rceil
\]
bisection tests. Since \(f\) is nondecreasing,
\[
    \sum_{k=2}^{K_T}
    \left(
        Q_k+s(k)
    \right)
    \le
    \sum_{k=2}^{K_T}
    \left(
        4+\left\lceil\log_2^+ B\right\rceil
        +\left\lceil f(k-1)\right\rceil
        +\left\lceil f(k)\right\rceil
    \right).
\]
Using \(\lceil f(k)\rceil\le f(k)+1\) and monotonicity,
\[
    \sum_{k=2}^{K_T}
    \left(
        Q_k+s(k)
    \right)
    \le
    C_BK_T+2K_Tf(K_T),
\]
with
\[
    C_B=6+\left\lceil\log_2^+B\right\rceil .
\]
\end{proof}
\begin{lemma}[Eventual bisection resolution]
\label{lem:endpoint_bisection_convergence}
Assume that
\[
    \frac{\ell_k-\ell_{k-1}}{\ell_k}\to0
\]
and that $f(k)\to\infty$. Then the endpoint-revalidated warm-started bisection
satisfies
\[
    \eta_k\to0.
\]
In particular, there exists a finite epoch $k_{\mathrm{bis}}$ such that
\[
    \eta_k<\frac{\Delta}{4},
    \qquad
    \forall k\ge k_{\mathrm{bis}}.
\]
\end{lemma}

\begin{proof}
By Lemma~\ref{lem:empirical_value_stability},
\[
    D_k\le B\frac{\ell_k-\ell_{k-1}}{\ell_k}\to0.
\]
Also \(2^{-f(k)}\to0\) and \(2^{1-s(k)}\to0\). Let
\(a_k=2^{1-s(k)}\). Fix \(\varepsilon>0\). For all sufficiently large \(k\),
\[
    2^{-f(k)}\le \varepsilon,
    \qquad
    a_k\le \frac14,
    \qquad
    a_kD_k\le \frac{\varepsilon}{4}.
\]
Using Lemma~\ref{lem:endpoint_revalidation_repair},
\[
    \eta_k
    \le
    \max\left\{
        \varepsilon,
        \frac14\eta_{k-1}+\frac{\varepsilon}{4}
    \right\}.
\]
Iterating gives \(\limsup_k\eta_k\le \varepsilon\). Since \(\varepsilon>0\)
was arbitrary, \(\eta_k\to0\).
\end{proof}
\subsection{Implementing the matching phase}
\label{subsec:dist_auction}
To complete the description of the algorithm, we need an algorithm that distributedly tests whether $G_{\Rh}(N,N,\gt)$ has a perfect matching or not, without messaging rewards to other agents.

For each threshold $\tau$, define the threshold graph
\[
E(\tau)=\{(n,m):\widehat R_{n,m}(k)\ge \tau\}.
\]
The feasibility test for $\tau$ is exactly the maximum-cardinality
bipartite matching problem on this graph. Algorithm~6 is a collision-only
implementation of the distributed auction/push-relabel matching procedure
analyzed in~\cite{naparstek2016expected}. The local prices in Algorithm~6
correspond to the local labels/prices in that procedure, and collisions
implement the reassignment and price-increment operations. Thus, the
matching subroutine is not a new matching algorithm; the novelty here is
its use as a privacy-preserving feasibility oracle inside the max-min
threshold bisection by using collision for the push operation.

 By Lemma 7 in \cite{naparstek2016expected} the fully distributed auction worst-case convergence time is $N^2(N-1)$ iterations. All agents are set to the unassigned state. They also set the price of arms with an estimated value $\gt$ or above to 0, and of other arms to $\infty$. An unassigned agent bids only if its minimum local price is finite; if all
local prices are infinite, then the agent has no incident edge in the threshold graph and remains unassigned. For $N^2(N-1)$ iterations, the following is performed: In iteration $iN+j$, if agent  $j$ is unassigned, it selects the arm with minimal price, accesses the arm, and assigns itself to the arm. All assigned agents access their arms. If an agent that was assigned to the arm experiences a collision, it becomes unassigned, and it increases the local price of the arm by 1.
 
 After $N^2(N-1)$ iterations of this algorithm, all unassigned agents sample all the arms to notify that the problem is infeasible and there is no perfect matching with all arm values $\gt$ or above. After experiencing the collision, all agents update $\gt_{\max}$. If no collision occurs in this period, all agents know that there is a feasible assignment and update $\gt_{\min}$. Agents save their assignment as their arm in the max-min assignment until a higher $\gt$ is proved feasible or the process terminates. 

 \begin{lemma}[Correctness and duration of one distributed feasibility test]
\label{lem:distributed_feasibility_test}
Fix an epoch \(k\) and a threshold \(\tau\). Let
\[
    G_{\tau}
    =
    G_{\widehat R(k)}(N,N,\tau)
\]
be the threshold graph with edge set
\[
    E(\tau)
    =
    \{(n,m):\widehat R_{n,m}(k)\ge \tau\}.
\]
One call to the distributed auction feasibility test terminates after at most
\[
    A_N \triangleq N^2(N-1)+N
\]
slots.

If \(G_{\tau}\) contains a perfect matching, then all agents output matched arms,
the resulting assignment is a perfect matching in \(G_{\tau}\), and no infeasibility signal is observed in the final notification schedule.

If \(G_{\tau}\) does not contain a perfect matching, then at least one agent
remains unmatched after the auction phase. During the final notification
schedule, every agent observes a collision, and therefore all agents output
\(\mathrm{infeasible}\).
\end{lemma}

\begin{proof}
The feasibility test consists of two parts. The first part is the distributed
auction phase. By the convergence bound for the distributed auction algorithm,
applied to the threshold graph \(G_{\tau}\), this phase terminates after at most
\(N^2(N-1)\) slots. The second part is the notification phase, which lasts
exactly \(N\) additional slots. Hence, the total duration is at most
\[
    N^2(N-1)+N=A_N .
\]

If \(G_{\tau}\) has a perfect matching, the distributed auction phase assigns
each agent to a distinct arm along an edge of \(G_{\tau}\). Thus, all agents are
matched at the end of the auction phase. During the final notification schedule,
each matched agent accesses only its assigned arm in the prescribed slot. Since
the assigned arms are distinct, no collision occurs. Therefore, no infeasibility
signal is observed, and the agents keep their matched arms as the feasible
assignment.

Conversely, suppose that \(G_{\tau}\) has no perfect matching. If all agents were
matched at the end of the auction phase, their assignments would form a perfect
matching in \(G_{\tau}\), contradicting infeasibility. Hence, at least one agent
remains unmatched. In the notification phase, each unmatched agent scans all
arms according to the common schedule. Every matched agent accesses its assigned
arm in the slot corresponding to that arm. Therefore, each matched agent collides
with an unmatched agent when the unmatched agent scans its assigned arm. Each
unmatched agent also observes a collision: if at least one agent is matched, the
unmatched agent collides when it scans a matched arm; if no agent is matched,
then, for \(N\ge2\), all unmatched agents follow the same notification scan and
collide with one another. The case \(N=1\) is trivial, since the single agent can
detect infeasibility locally from the absence of any feasible edge.

Thus, when \(G_{\tau}\) has no perfect matching, the final notification schedule
produces a common infeasibility signal observed by all agents. Therefore all
agents output \(\mathrm{infeasible}\).
\end{proof}

\subsection{Exploitation}
During the exploitation phase, each agent uses the last feasible allocation observed during the matching phase. The length of the phase during epoch $k$ is $c_{3,n}(k)=c^k$. While for the proof of the theorem we assume $c=2$, it is a meta-parameter which can assist in trading exploration and exploitation.

\section{Regret analysis}
\label{sec:regret_analysis}

We now prove the main regret bound. Throughout this section, let
\[
    B\ge \max_{n,m}B_{n,m}
\]
be a deterministic upper bound on the reward supports. This constant is used only
in the analysis, the agents do not need to know it.

\begin{theorem}[Explicit regret bound]
\label{thm:main}
Assume $N=M$, rewards are nonnegative and bounded, and the max-min gap satisfies
$\Delta>0$. Let $f(k)$ be nondecreasing\footnote{Without loss of generality we also assume that $f(k)\ge1$ for all $k\ge1$; otherwise replace $f$ by
$\max\{1,f\}$, which changes only the finite constant.} with $f(k)\to\infty$ and
$f(k-1)/f(k)\to 1$.
Equivalently, 
\[
    \frac{\ell_k-\ell_{k-1}}{\ell_k}\to0,
    \qquad
    \ell_k=Lk f(k).
\]
Let
\[
    K_T\triangleq \left\lceil \log_2(T+1)\right\rceil .
\]
Let the matching phase
use
\[
    s(k)=\lceil f(k)\rceil
\]
feasibility tests after endpoint revalidation and local repair. Then Algorithm~\ref{alg:FairBandits} satisfies
\[
    \Reg(T)
    \le
    C_0
    +
    BNLK_Tf(K_T)
    +
    C_{\mathrm m}(B)BN^3K_Tf(K_T),
\]
where \(C_0<\infty\) is independent of \(T\), and \(C_{\mathrm m}(B)\) is a
finite constant independent of \(T\).
Consequently,
\[
    \Reg(T)
    =
    O_{B,\Delta,L}
    \left(
        N^3\log T\, f(\log T)
    \right),
    \qquad N=M.
\]
\end{theorem}

\subsection{Exploration errors}

The estimates are cumulative across epochs. Let
\[
    \ell_k=Lk f(k)
\]
be the total number of samples collected from each pair $(n,m)$ up to the end of
epoch $k$.

\begin{lemma}[Exploration mean estimation error]
\label{lem:exploration_error}
For any $\gd>0$ and any epoch $k$,
\begin{equation}
\label{eq:exploration_error_lemma}
    \bbP\!\left(
        \exists\, n,m :
        \left|\Rh_{n,m}(k)-R_{n,m}\right|>\gd
    \right)
    \le
    2NM \exp\!\left(
        -\frac{2\ell_k\gd^2}{B^2}
    \right),
    \qquad
    \ell_k=Lk f(k).
\end{equation}
\end{lemma}

\begin{proof}
For each fixed pair $(n,m)$, the empirical mean $\Rh_{n,m}(k)$ is computed from
$\ell_k$ independent samples supported in $[0,B]$. Hoeffding's inequality gives
\[
    \bbP\!\left(
        |\Rh_{n,m}(k)-R_{n,m}|>\gd
    \right)
    \le
    2\exp\!\left(
        -\frac{2\ell_k\gd^2}{B^2}
    \right).
\]
A union bound over all $NM$ pairs proves the claim.
\end{proof}

\subsection{Probability of matching error}

We now relate the bisection output for the empirical rewards to true max-min optimality.

Let $k_{\mathrm{bis}}$ be the finite epoch guaranteed by
Lemma~\ref{lem:endpoint_bisection_convergence}, so that
\[
    \tau_{\max}(k)-\tau_{\min}(k)<\Delta/4,
    \qquad
    \forall k\ge k_{\mathrm{bis}}.
\]
\begin{lemma}[Eventual true optimality of the bisection output]
\label{lem:approx_matching}
For any assignment $\pi$, define
\[
    \rho(\pi)\triangleq \min_n R_{n,\pi(n)},
    \qquad
    \hat\rho_k(\pi)\triangleq \min_n \Rh_{n,\pi(n)}(k).
\]
Let
\[
    \rho^*\triangleq \max_{\pi\in S_N}\rho(\pi),
    \qquad
    A_1\triangleq \{\pi\in S_N:\rho(\pi)=\rho^*\},
\]
and define
\[
    \rho_2\triangleq \max_{\pi\notin A_1}\rho(\pi),
    \qquad
    \gD\triangleq \rho^*-\rho_2>0.
\]
At epoch $k$, let $\hat\pi(k)$ be the assignment returned by the bisection
procedure. If $k\ge k_{\mathrm{bis}}$ and
\[
    \max_{n,m}\left|\Rh_{n,m}(k)-R_{n,m}\right|<\frac{\gD}{4},
\]
then
\[
    \hat\pi(k)\in A_1.
\]
\end{lemma}

\begin{proof}
For $k\ge k_{\mathrm{bis}}$, the bisection resolution satisfies
\[
    \eta_k\triangleq \tau_{\max}(k)-\tau_{\min}(k)<\frac{\gD}{4}.
\]
Thus the bisection output is $\eta_k$-optimal for the empirical max-min problem:
\[
    \hat\rho_k(\hat\pi(k))
    \ge
    \max_{\pi\in S_N}\hat\rho_k(\pi)-\eta_k.
\]
Let $\pi^*\in A_1$. If the uniform estimation error is less than $\gD/4$, then
\[
    \max_{\pi\in S_N}\hat\rho_k(\pi)
    \ge
    \hat\rho_k(\pi^*)
    >
    \rho^*-\frac{\gD}{4}.
\]
Therefore,
\[
    \hat\rho_k(\hat\pi(k))
    >
    \rho^*-\frac{\gD}{4}-\frac{\gD}{4}
    =
    \rho^*-\frac{\gD}{2}.
\]
Suppose, toward a contradiction, that $\hat\pi(k)\notin A_1$. Then
$\rho(\hat\pi(k))\le \rho_2=\rho^*-\gD$. Using the uniform estimation bound again,
\[
    \hat\rho_k(\hat\pi(k))
    <
    \rho(\hat\pi(k))+\frac{\gD}{4}
    \le
    \rho^*-\frac{3\gD}{4},
\]
contradicting the previous lower bound. Hence $\hat\pi(k)\in A_1$.
\end{proof}

Combining Lemma~\ref{lem:approx_matching} with
Lemma~\ref{lem:exploration_error}, and taking $\gd=\gD/4$, gives for every
$k\ge k_{\mathrm{bis}}$,
\begin{equation}
\label{eq:bisection_failure_probability}
    \bbP\!\left(\hat\pi(k)\notin A_1\right)
    \le
    2NM\exp\!\left(-a_{\Delta}Lk f(k)\right),
    \qquad
    a_{\Delta}=\frac{\gD^2}{8B^2}.
\end{equation}

\subsection{Regret computation}

We now compute the regret contribution of each phase.

\paragraph{Exploration regret.}
In epoch $k$, the exploration phase has length
\[
    c_1(k)=ML\left(g(k)-g(k-1)\right),
    \qquad
    g(k)=k f(k).
\]
Therefore, by telescoping,
\[
    \sum_{k=1}^{K_T}c_1(k)=MLK_T f(K_T).
\]
Using the crude per-sample bound $B$,
\begin{equation}
\label{eq:reg1}
    \Reg_{\mathrm{explore}}(T)
    \le
    BM L K_T f(K_T).
\end{equation}
Since $N=M$,
\begin{equation}
\label{eq:reg1_NM}
    \Reg_{\mathrm{explore}}(T)
    \le
    BNL K_T f(K_T).
\end{equation}

\paragraph{Ordering and matching regret.}
The initial ordering phase contributes at most
\[
    BT_{\mathrm{ord}}
\]
regret. The one-time initial bracketing stage contributes at most
\[
    BA_N b_B
\]
regret. These two finite terms are included in $C_0$.

In each epoch, the matching phase performs two endpoint-revalidation tests,
possibly a local geometric repair, and then $s(k)=\lceil f(k)\rceil$ bisection
tests. By Lemma~\ref{lem:endpoint_revalidation_repair}, the number of local
repair tests is controlled by the ratio between the empirical drift and the
stored bracket width. With the public padding rule $w_k=2^{-f(k)}$.
By Lemma~\ref{lem:geometric_repair_cost}, the total number of distributed
feasibility tests used for endpoint revalidation, geometric repair, and
bisection over the first \(K_T\) epochs is at most
\[
    C_BK_T+2K_Tf(K_T).
\]
Each feasibility test takes at most
\[
    A_N=N^2(N-1)+N
\]
slots by Lemma~\ref{lem:distributed_feasibility_test}. Therefore, the regret contribution of the matching phases is bounded by
\[
    \Reg_{\mathrm{match}}(T)
    \le
    BA_N\left(C_BK_T+2K_Tf(K_T)\right),
\]
up to the finite initial ordering and initial bracketing terms.
which are absorbed into $C_0$. Therefore, for an absolute constant
$C_{\mathrm{m}}$,
\[
    \Reg_{\mathrm{match}}(T)
    \le
    BT_{\mathrm{ord}}
    +
    BA_N b_B
    +
    C_{\mathrm{m}}BA_N
    \sum_{k=1}^{K_T} f(k).
\]
Since $f$ is nondecreasing,
\[
    \sum_{k=1}^{K_T}f(k)
    \le
    K_T f(K_T).
\]
Thus, using $A_N\le N^3$,
\[
    \Reg_{\mathrm{match}}(T)
    \le
    BT_{\mathrm{ord}}
    +
    BA_N b_B
    +
    C_{\mathrm{m}}BN^3K_T f(K_T).
\]

\paragraph{Exploitation regret.}
For epochs $k<k_{\mathrm{bis}}$, we use the crude bound that every exploitation
slot may incur regret at most $B$. For epochs $k\ge k_{\mathrm{bis}}$,
Lemma~\ref{lem:approx_matching} shows that the exploitation allocation can be
incorrect only if the empirical rewards are not sufficiently accurate. Therefore, by
\eqref{eq:bisection_failure_probability},
\[
    \bbP(\text{exploitation error in epoch }k)
    \le
    2NM\exp\!\left(-a_{\Delta}Lk f(k)\right).
\]
For \(a>0\) and \(k_0\ge1\), define
\[
    S_f(a,L,k_0)
    \triangleq
    \sum_{k=k_0}^{\infty}
        2^k\exp(-aLkf(k)).
\]
Since \(f(k)\to\infty\), this series is finite.
Since the exploitation phase in epoch $k$ has length $2^k$,
\begin{align}
\label{eq:reg3}
    \Reg_{\mathrm{exploit}}(T)
    &\le
    B\sum_{k=1}^{k_{\mathrm{bis}}-1}2^k 
    +
    B\sum_{k=k_{\mathrm{bis}}}^{K_T}2^k
    \left[
        2NM\exp\!\left(-a_{\Delta}Lk f(k)\right)
    \right] \nonumber\\
    &\le
    B\left(2^{k_{\mathrm{bis}}}-2\right)
    +
    BNM\left[
        2S_f(a_{\Delta},L,k_{\mathrm{bis}})
    \right].
\end{align}
The last term is finite because $f(k)\to\infty$, equivalently because the total
number of samples per pair, $\ell_k=Lk f(k)$, is superlinear in $k$.

Combining the exploration, matching, and exploitation bounds, and grouping the
finite terms into \(C_0\), yields
\[
    \Reg(T)
    \le
    C_0
    +
    BNLK_Tf(K_T)
    +
    C_{\mathrm m}BN^3K_Tf(K_T).
\]
Since $K_T=\lceil \log_2(T+1)\rceil$ and $f$ is nondecreasing,
\[
    \Reg(T)
    =
    O_{B,\gD,L}\left(N^3\log T\,f(\log T)\right),
    \qquad N=M.
\]
This proves Theorem~\ref{thm:main}.

\subsection{Concrete schedules and finite constants}
\label{subsec:concrete_schedules}

We now instantiate Theorem~\ref{thm:main} for two concrete schedules. The first
choice gives the near-logarithmic regret rate, while the second gives a smaller
finite bisection transient.

Throughout this subsection, let
\[
    A_N\triangleq N^2(N-1)+N
\]
be the number of slots required by one distributed feasibility test, and let
\[
    q_B\triangleq \max\{0,\lceil\log_2 B\rceil+1\},
    \qquad
    \bar B\triangleq 2^{q_B},
    \qquad
    b_B\triangleq q_B+1 .
\]
The quantity \(b_B\) bounds the number of distributed feasibility tests in the
one-time initial bracketing stage.

For a given schedule \(f\), define
\[
    \ell_k=Lk f(k),
    \qquad
    \bar D_k
    \triangleq
    B\frac{\ell_k-\ell_{k-1}}{\ell_k}.
\]
By Lemma~\ref{lem:empirical_value_stability}, \(\bar D_k\) is a deterministic
upper bound on the change of the empirical max-min value between epochs
\(k-1\) and \(k\).

Define the deterministic bracket-width envelope by
\[
    \bar\eta_1=\bar B,
\]
and, for \(k\ge2\),
\[
    \bar\eta_k
    =
    \max\left\{
        2^{-f(k)},
        2^{1-\lceil f(k)\rceil}
        \left(
            \bar\eta_{k-1}+\bar D_k
        \right)
    \right\}.
\]
The first term is the public padding width stored after bisection. The second
term follows from endpoint revalidation, local repair, and the subsequent
\(\lceil f(k)\rceil\) bisection tests.

We define
\[
    k_{\mathrm{bis}}
    \triangleq
    \inf\left\{
        k\ge1:
        \bar\eta_j<\frac{\Delta}{4}
        \text{ for all } j\ge k
    \right\}.
\]
Since \(f(k)\to\infty\) and \((\ell_k-\ell_{k-1})/\ell_k\to0\), we have
\(\bar\eta_k\to0\). Hence \(k_{\mathrm{bis}}<\infty\), and it is independent of
the horizon $T$.
Let $a_\Delta
    \triangleq
    \frac{\Delta^2}{8B^2}$,
For $a>0$, define
\[
    S_f(a,L,k_{\mathrm{bis}})
    \triangleq
    \sum_{k=k_{\mathrm{bis}}}^{\infty}
        2^k\exp(-aLk f(k)).
\]
This series is finite because \(f(k)\to\infty\). Therefore, for \(N=M\), the
constant in Theorem~\ref{thm:main} can be written as
\begin{equation}
\label{eq:C0_concrete_general}
    C_0
    =
    BT_{\mathrm{ord}}
    +
    BA_Nb_B
    +
    B(2^{k_{\mathrm{bis}}}-2)
    +
    2BN^2S_f(a_\Delta,L,k_{\mathrm{bis}}).
\end{equation}

\paragraph{Near-logarithmic schedule.}
Consider
\[
    f(k)=r\log_2(k+1),
    \qquad r\ge1.
\]
Then
\[
    \ell_k=Lrk\log_2(k+1),
\]
and, for \(k\ge2\),
\[
    \bar D_k
    =
    B
    \frac{
        k\log_2(k+1)-(k-1)\log_2 k
    }{
        k\log_2(k+1)
    }
    =
    O\!\left(\frac{B}{k}\right).
\]
Moreover,
\[
    2^{-f(k)}
    =
    (k+1)^{-r}.
\]
Thus, up to universal constants, the bisection transient satisfies
\[
    k_{\mathrm{bis}}
    =
    O\!\left(
        \left(
            \frac{\bar B+B}{\Delta}
        \right)^{1/r}
    \right).
\]
For this schedule,
\[
    S_f(a,L,k_{\mathrm{bis}})
    =
    \sum_{k=k_{\mathrm{bis}}}^{\infty}
        2^k
        \exp\!\left(
            -aLrk\log_2(k+1)
        \right).
\]
The regret bound becomes
\[
    \Reg(T)
    =
    O_{B,\Delta,L,r}
    \left(
        N^3\log T\,\log\log T
    \right),
    \qquad N=M.
\]

\paragraph{Alternative schedule.}
To show the dependence of $C_0$ on the parameter we also 
consider
\[
    f(k)=k^\epsilon,
    \qquad 0<\epsilon<1.
\]
Then
\[
    \ell_k=Lk^{1+\epsilon},
\]
and
\[
    \bar D_k
    =
    B
    \left[
        1-\left(1-\frac1k\right)^{1+\epsilon}
    \right]
    \le
    \frac{B(1+\epsilon)}{k}.
\]
Moreover,
\[
    2^{-f(k)}
    =
    2^{-k^\epsilon}.
\]
Thus, up to universal constants,
\[
    k_{\mathrm{bis}}
    =
    O\!\left(
        \left[
            \log_2\!\left(
                \frac{\bar B+B(1+\epsilon)}{\Delta}
            \right)
        \right]^{1/\epsilon}
    \right).
\]
For this schedule,
\[
    S_f(a,L,k_{\mathrm{bis}})
    =
    \sum_{k=k_{\mathrm{bis}}}^{\infty}
        2^k
        \exp(-aLk^{1+\epsilon}).
\]
The regret bound becomes
\[
    \Reg(T)
    =
    O_{B,\Delta,L,\epsilon}
    \left(
        N^3(\log T)^{1+\epsilon}
    \right),
    \qquad N=M.
\]

The schedule \(f(k)=r\log_2(k+1)\) gives the sharper asymptotic dependence on
\(T\). The schedule \(f(k)=k^\epsilon\) gives a smaller bisection transient bound. Thus,
the choice of \(f\) provides a tunable tradeoff between asymptotic regret and
finite-time constants.

\paragraph{Remark on the finite constant.}
The explicit constant \(C_0\) is a worst-case transient bound and is not expected
to be tight. In particular, the term \(B(2^{k_{\mathrm{bis}}}-2)\) pessimistically
charges every exploitation slot before the asymptotic regime as if it incurred
maximal regret. In practice, the empirical allocation is often optimal well
before the epoch \(k_{\mathrm{bis}}\) guaranteed by the proof, and even non-optimal
allocations may have max-min value close to \(\rho^*\). Consequently, the
finite-time constant observed in simulations is much smaller than the conservative
bound used in the theorem.
\section{Simulations}

 \subsection{Simulations for various number of agents}
\label{appendix_simulations}
To demonstrate the scalability of the algorithm, we present 1000 Monte-Carlo experiments for $N=2,4,8,16, 32, 64  \hbox{\ and \ } 128$ agents. The minimal gap was $\frac{1}{N}$, by selecting the arm values for each agent as a random permutation of the numbers $\frac{k}{N}: k=1,\ldots, N$. The number of epochs was set to 32. We can clearly see the logarithmic dependence of the cumulative regret on $T$ for each value of $N$, as well as the polynomial dependence on the number of agents.

\begin{figure}
  \centering
  \includegraphics[width = 0.8\columnwidth]{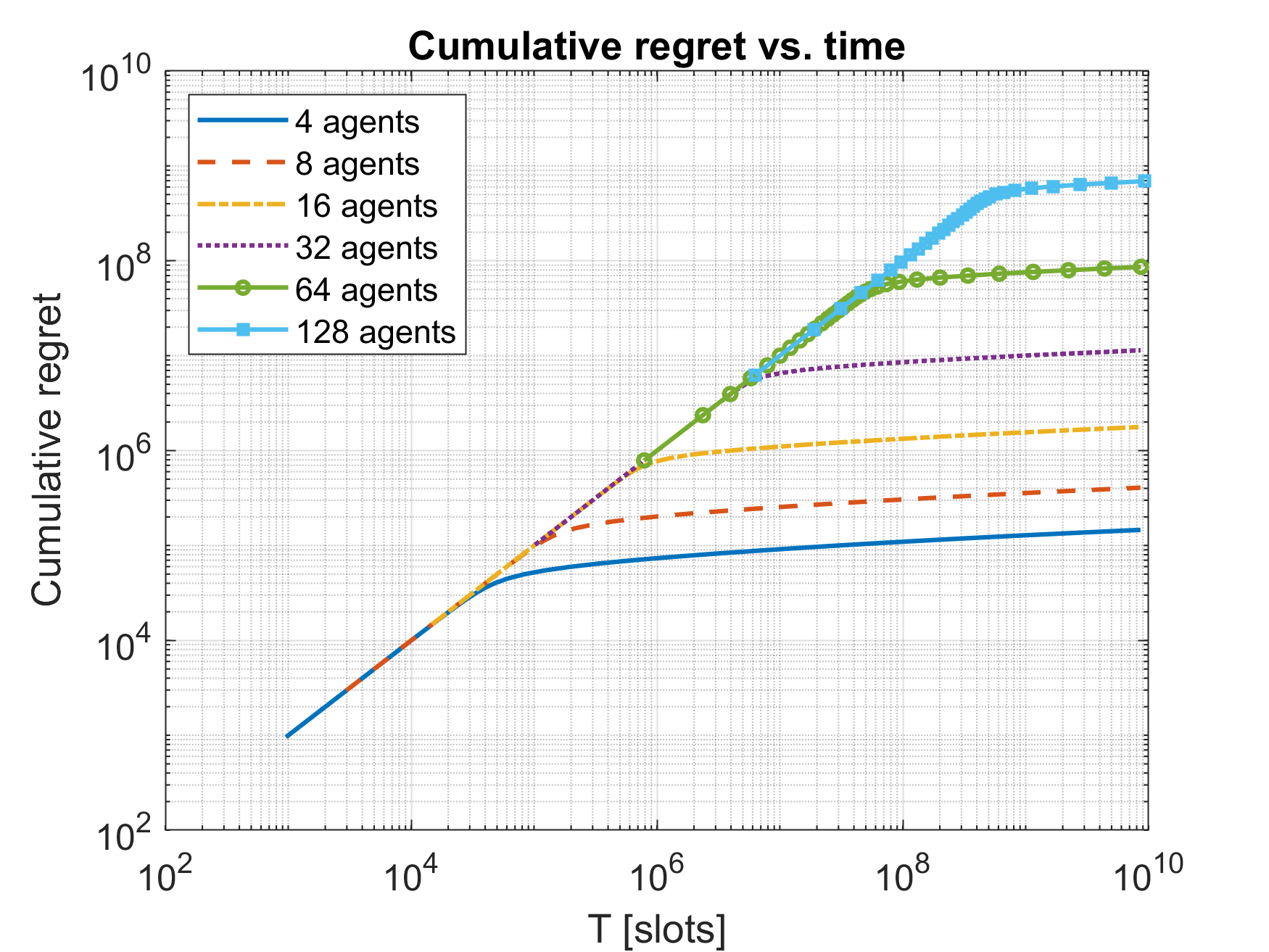}
\caption{Cumulative regret as a function of time for the proposed algorithm.
The plot shows the empirical median cumulative regret over 1000 independent
Monte-Carlo trials for $N=M\in\{4,8,16,32,64,128\}$ agents/arms.  The reward
matrix in each trial is generated by assigning each agent a random permutation
of the values $k/N$, $k=1,\ldots,N$, so the minimum max-min gap is $1/N$.
The experiment uses 32 epochs, and both axes are shown on logarithmic scale.}
\label{fig:cumulative_regret_T1}
\end{figure}

Figure \ref{fig:cumulative_regret_T1} presents the median regret over $1000$ Monte-Carlo tests as a function of time on a logarithmic scale. We can clearly see the near-linear growth with $\log T$ for various numbers of agents.  
Figure 3 presents the median regret at several epoch endpoints as a function of the number of agents over 1000 Monte-Carlo trials. The observed growth is consistent with the predicted polynomial dependence on $N$.
\begin{figure}
  \centering
  \includegraphics[width = 0.8\columnwidth]{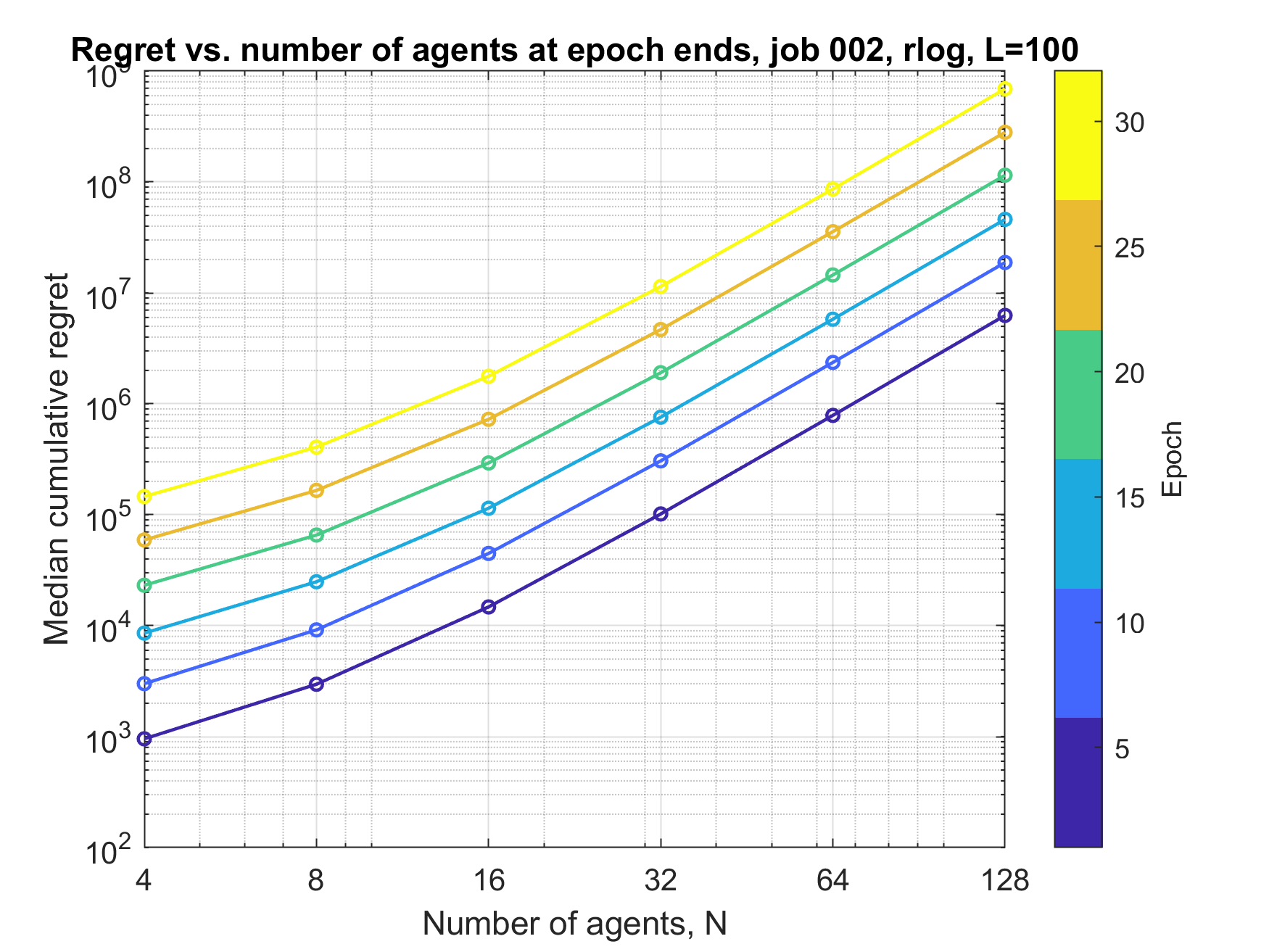}
\caption{Median cumulative regret as a function of the number of agents.
The plot shows 1000-trial empirical medians for
$N=M\in\{4,8,16,32,64,128\}$ agents/arms.  Each curve corresponds to a fixed
epoch endpoint, with the epoch index indicated by the color bar.  The reward
matrices are generated as in Fig.~2, with each agent assigned a random
permutation of $k/N$, $k=1,\ldots,N$, giving minimum max-min gap $1/N$.
Both axes are logarithmic.}
\label{fig:cumulative_regret_N1}
\end{figure}
\paragraph{Dependence on the max-min gap.}
We next examine the dependence of the regret on the max-min gap $\Delta$.
For this experiment we fixed $N=8$, used $L=100$, and ran $10{,}000$ independent
Monte-Carlo trials for each value of $\Delta$. The reward matrices were generated
with a controlled max-min gap, so that the optimal max-min assignment and the best non-optimal assignment differ by the prescribed value of $\Delta$.
Figure~\ref{fig:regret_delta} reports the final cumulative regret after $32$
epochs as a function of $1/\Delta$. The curves show the empirical $1\%$ best
case, the median, and the empirical $1\%$ worst case over the $10{,}000$ trials.
These empirical quantiles display the tail behavior of the algorithm, with
$10{,}000$ trials, the $1\%$ tail corresponds to approximately $100$ runs.

In the tested range, the median regret is relatively insensitive to $1/\Delta$,
indicating that the dominant cost in this finite-horizon regime is the fixed
exploration and matching overhead rather than persistent exploitation of
incorrect allocations. The upper empirical tail increases for smaller gaps, as
expected, since a smaller $\Delta$ makes it harder to distinguish the optimal
max-min assignment from near-optimal alternatives. This trend is consistent
with the concentration term in the analysis, where the exponent scales with
$\Delta^2$.
\paragraph{Probability of non-optimal exploitation allocation.}
To directly illustrate the concentration behavior behind Lemma~\ref{lem:approx_matching},
we also measured the probability that the allocation used in the exploitation phase is
not max-min optimal. We fixed \(N=8\), \(L=100\), and ran \(10{,}000\) independent
Monte-Carlo trials for each value of the max-min gap
\[
    \Delta\in\{0.25,0.125,0.0625,0.03125\}.
\]
Figure~\ref{fig:nonoptimal_epoch_probability} reports the empirical fraction of
non-optimal exploitation allocations as a function of the epoch. The decay is
monotone and becomes faster as the gap increases, as predicted by the concentration
term \(\exp(-a_\Delta Lk f(k))\), where \(a_\Delta=\Delta^2/(8B^2)\). For the
larger gaps, the empirical error probability falls below \(10^{-3}\) within only
a few epochs. For the smallest tested gap, \(\Delta=0.03125\), the decay is slower
but still reaches the \(10^{-3}\) level by roughly epoch \(24\). Since the experiment
uses \(10{,}000\) trials, probabilities near \(10^{-3}\) correspond to about ten
observed error events, so the plot captures the relevant rare-event regime for
the exploitation-error analysis.
The main reason the parameter $\Delta$ impacts the performance is that a smaller $\Delta$, increases convergence time, and therefore, for a given epoch, increases the probability of a sub-optimal allocation. 

\begin{figure}
  \centering
  \includegraphics[width=0.8\textwidth]{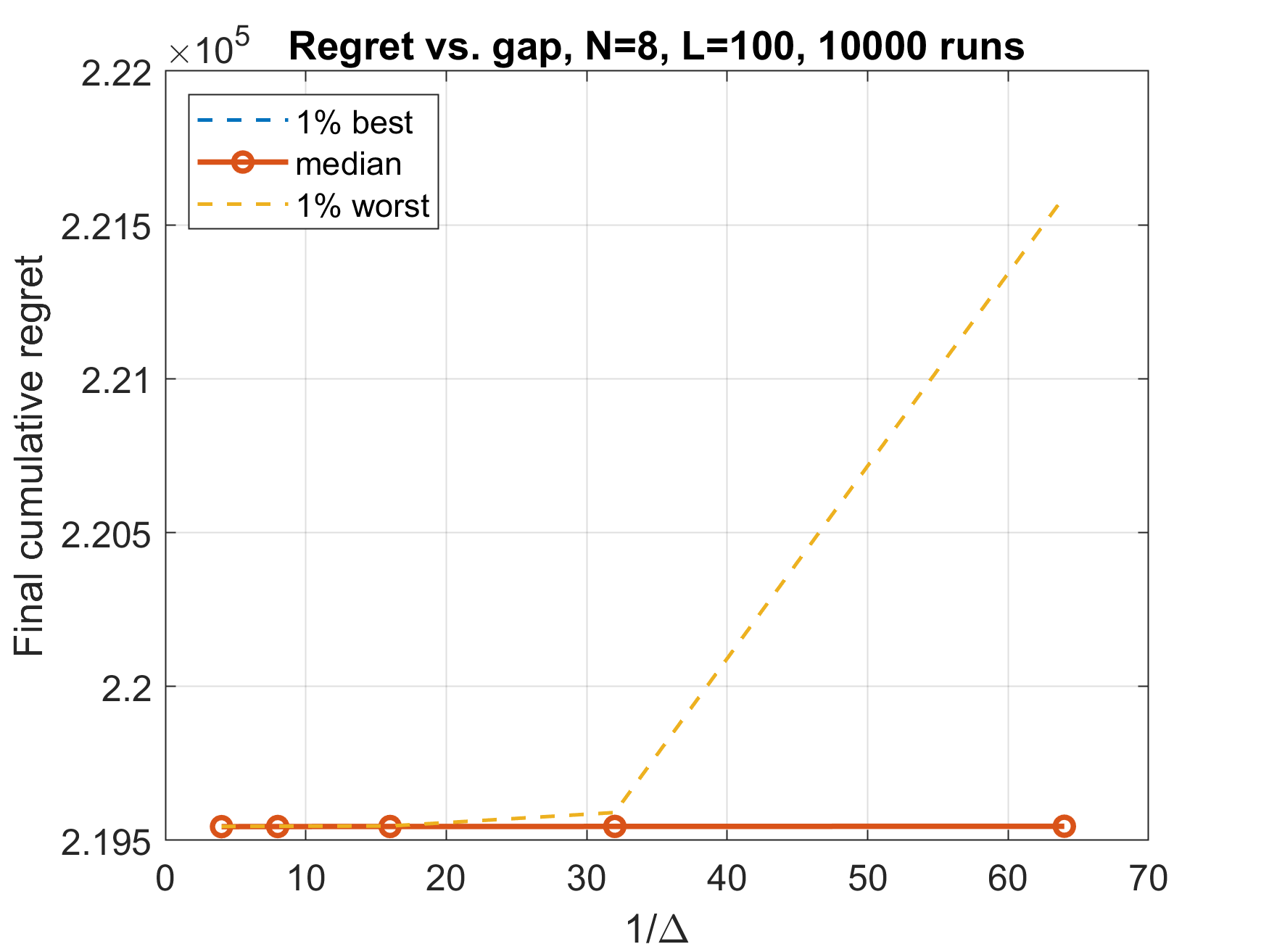}
  \caption{Final cumulative regret as a function of $1/\Delta$ for $N=8$,
  $L=100$, and $10000$ Monte-Carlo trials. The curves show the empirical
  $1\%$ best case, median, and empirical $1\%$ worst case.}
  \label{fig:regret_delta}
\end{figure}

\begin{figure}
  \centering
  \includegraphics[width=0.8\textwidth]{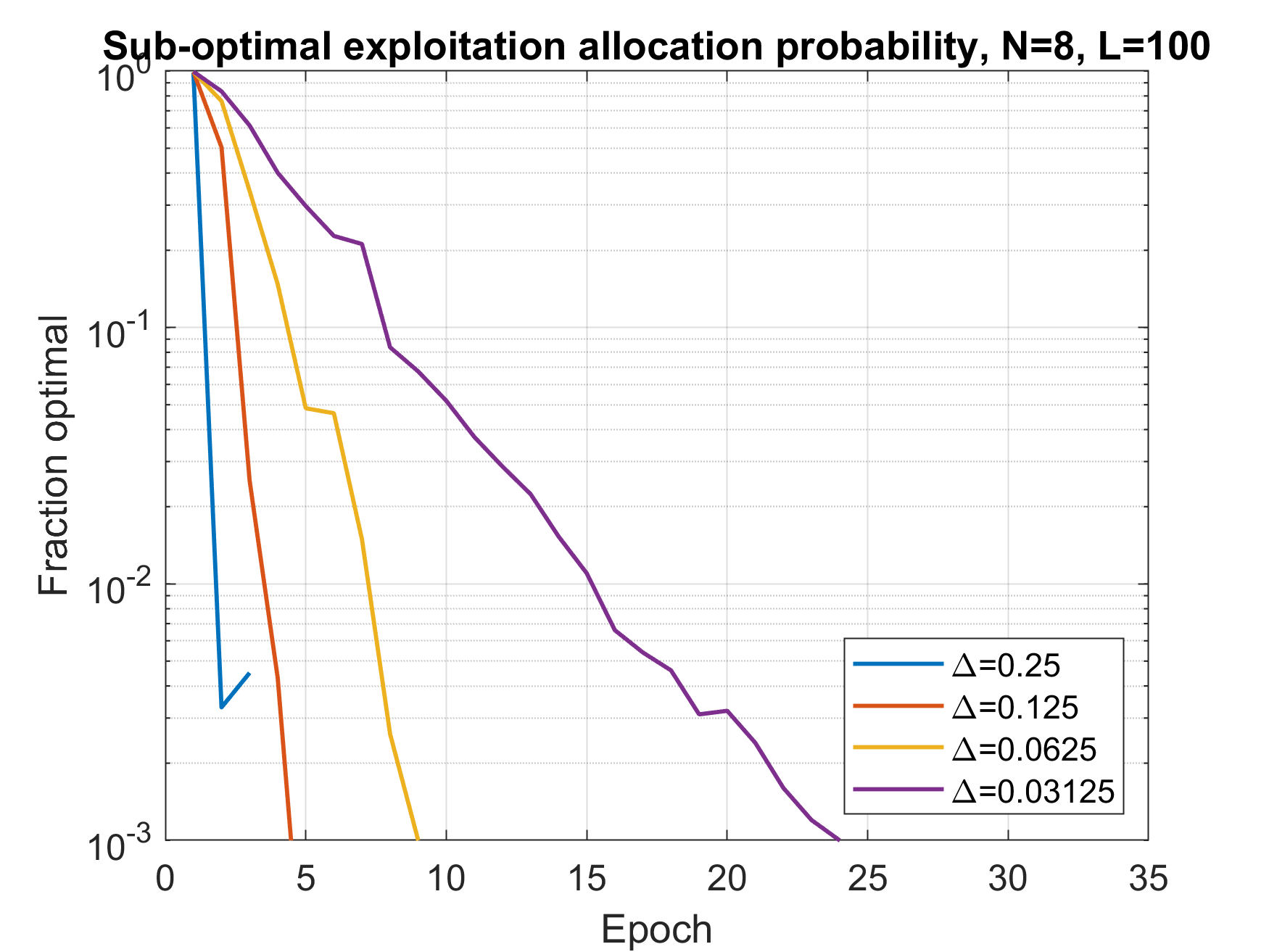}
  \caption{Empirical probability of a non-optimal exploitation allocation as a
  function of the epoch. The experiment uses \(N=8\), \(L=100\), and \(10{,}000\)
  Monte-Carlo trials for each gap value
  \(\Delta\in\{0.25,0.125,0.0625,0.03125\}\). The curves show that the probability
  of selecting a non-optimal max-min allocation decays rapidly with the epoch,
  and that the decay is faster for larger max-min gaps. With \(10{,}000\) trials,
  the \(10^{-3}\) level corresponds to approximately ten error events.}
  \label{fig:nonoptimal_epoch_probability}
\end{figure}

\begin{figure}
  \centering
  \includegraphics[width=0.8\textwidth]{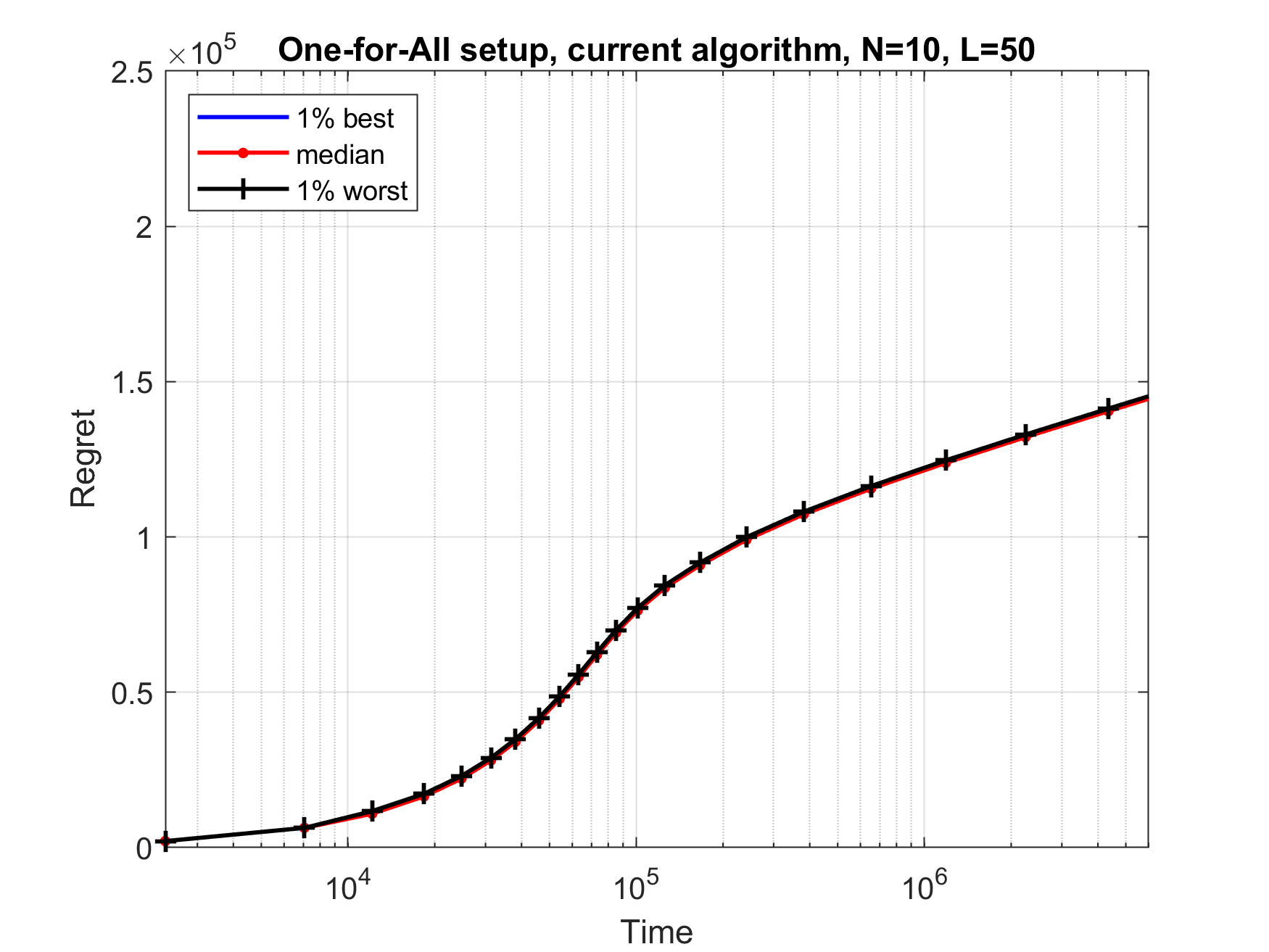}
  \caption{Comparison with the setup of \cite{bistritz2021one}. The proposed
  endpoint-revalidated algorithm is run with $N=10$, $L=50$, and exploitation
  length $c_3(k)=2^k$. The curves show empirical $1\%$ best, median, and
  empirical $1\%$ worst regret over $1000$ trials, overlapping with each other.}
  \label{fig:compare_one_forall}
\end{figure}
\paragraph{Comparison with One-for-All and All-for-One.}
Finally, we compare the proposed algorithm with the fair-bandit algorithm of
\cite{bistritz2021one} using the same ten-agent instance and Gaussian reward
noise considered there. The regret bound in Theorem~\ref{thm:main} is proved for
bounded rewards, so this experiment is outside the formal assumptions of the
theorem; nevertheless, it is useful for comparison with the prior benchmark. We
used the endpoint-revalidated version of the proposed algorithm with $L=50$ and
exploitation length $c_3(k)=2^k$, and ran $1000$ independent trials. The figure
shows the empirical $1\%$ best case, median, and empirical $1\%$ worst case. In
the benchmark reported in \cite{bistritz2021one}, the median regret at
$T=6\times 10^6$ was approximately $5\times10^5$ over $100$ experiments. In our
simulation, the median regret at the same horizon is about $1.4\times10^5$, and
the empirical $1\%$ worst-case curve remains below $1.5\times10^5$. Thus, on
this instance, the proposed distributed matching approach reduces regret by a
factor of roughly three to four relative to the previously reported benchmark,
while requiring only collision-based coordination and no sharing of reward
samples or empirical reward estimates.
\section{Conclusion}
Fair allocation of resources is an important objective in decentralized learning systems, especially when agents have heterogeneous utilities and limited ability
or willingness to share information. In this paper, we studied the max-min fair multi-agent multi-armed bandit problem under collision-only coordination. We
proposed a fully distributed algorithm that achieves near-logarithmic regret in the horizon and polynomial dependence on the number of agents. The algorithm
combines distributed agent ordering, fixed-length cumulative exploration, endpoint-revalidated bisection, and collision-based distributed auction tests for threshold feasibility. In contrast to leader-based or coded communication
approaches, no agent needs to collect or reconstruct another agent's rewards or
empirical utility estimates.

The results show that tools from distributed computation can substantially improve the scalability of fair bandit learning. In particular, the distributed auction subroutine replaces the exponentially large Markov-chain matching
process used in prior fair-bandit algorithms by a polynomial-time feasibility test. The simulations support the theoretical findings, showing favorable scaling with the horizon, the number of agents, and the max-min gap, as well as
improved performance relative to the previously proposed One-for-All and All-for-One algorithm.

Several extensions are natural. First, the assumption \(N=M\) is mainly for notational simplicity. When there are more arms than agents, say $M>N$, the
threshold feasibility test only needs to find a matching that saturates all
agents. Equivalently, one may add \(M-N\) dummy agents with universal edges; the
dummy agents only fill unused arms and are not included in the max-min
objective. The distributed auction subroutine then applies to a rectangular
\(N\times M\) matching problem, with worst-case feasibility-test complexity
\(O(N^2M)\). Thus, the regret bounds extend by replacing the \(N^3\) matching
factor by \(N^2M\). When there are more agents than arms, the arm-sharing
techniques of~\cite{boyarski2023distributed} may be useful, although combining
arm sharing with max-min fair learning remains an interesting direction.

A second extension is weighted max-min fairness. Given positive weights, each
agent can locally rescale its empirical rewards before the threshold tests. For
example, under the objective
\[
    \max_{\pi}\min_{n\in[N]} w_n R_{n,\pi(n)},
\]
agent \(n\) simply replaces the threshold condition
\(\widehat R_{n,m}(k)\ge \tau\) by
\[
    w_n\widehat R_{n,m}(k)\ge \tau .
\]
The same endpoint-revalidated bisection and distributed auction procedures then
apply without modification. This yields a family of Pareto-dominant allocations
in which the achieved rewards need not be equally balanced; instead, the weights
determine the relative priority or service levels of the agents.

The main open question is whether exact logarithmic regret can be achieved
under the same collision-only and no-reward-sharing constraints. Another important direction is extending the analysis beyond bounded rewards, for
example to sub-Gaussian reward distributions, while preserving a simple
polynomial bound on the distributed matching phase.
\section*{Acknowledgement} ChatGPT was utilized to generate sections of this work, including text, tables, code, and citations based on an original text written by the author. It was also used to review the manuscript and debug the algorithm presented.
\newpage


\begin{thebibliography}{53}


\ifx \showCODEN    \undefined \def \showCODEN     #1{\unskip}     \fi
\ifx \showISBNx    \undefined \def \showISBNx     #1{\unskip}     \fi
\ifx \showISBNxiii \undefined \def \showISBNxiii  #1{\unskip}     \fi
\ifx \showISSN     \undefined \def \showISSN      #1{\unskip}     \fi
\ifx \showLCCN     \undefined \def \showLCCN      #1{\unskip}     \fi
\ifx \shownote     \undefined \def \shownote      #1{#1}          \fi
\ifx \showarticletitle \undefined \def \showarticletitle #1{#1}   \fi
\ifx \showURL      \undefined \def \showURL       {\relax}        \fi
\providecommand\bibfield[2]{#2}
\providecommand\bibinfo[2]{#2}
\providecommand\natexlab[1]{#1}
\providecommand\showeprint[2][]{arXiv:#2}

\bibitem[Alatur et~al\mbox{.}(2020)]%
        {alatur2020multi}
\bibfield{author}{\bibinfo{person}{Pragnya Alatur}, \bibinfo{person}{Kfir~Y Levy}, {and} \bibinfo{person}{Andreas Krause}.} \bibinfo{year}{2020}\natexlab{}.
\newblock \showarticletitle{Multi-Player Bandits: The Adversarial Case}.
\newblock \bibinfo{journal}{\emph{J. Mach. Learn. Res.}} \bibinfo{volume}{21}, \bibinfo{number}{77} (\bibinfo{year}{2020}), \bibinfo{pages}{1--23}.
\newblock


\bibitem[Anandkumar et~al\mbox{.}(2011)]%
        {Anandkumar2011}
\bibfield{author}{\bibinfo{person}{Animashree Anandkumar}, \bibinfo{person}{Nithin Michael}, \bibinfo{person}{Ao~Kevin Tang}, {and} \bibinfo{person}{Ananthram Swami}.} \bibinfo{year}{2011}\natexlab{}.
\newblock \showarticletitle{Distributed algorithms for learning and cognitive medium access with logarithmic regret}.
\newblock \bibinfo{journal}{\emph{IEEE Journal on Selected Areas in Communications}} \bibinfo{volume}{29}, \bibinfo{number}{4} (\bibinfo{year}{2011}), \bibinfo{pages}{731--745}.
\newblock


\bibitem[Asadpour and Saberi(2010)]%
        {asadpour2010approximation}
\bibfield{author}{\bibinfo{person}{Arash Asadpour} {and} \bibinfo{person}{Amin Saberi}.} \bibinfo{year}{2010}\natexlab{}.
\newblock \showarticletitle{An approximation algorithm for max-min fair allocation of indivisible goods}.
\newblock \bibinfo{journal}{\emph{SIAM J. Comput.}} \bibinfo{volume}{39}, \bibinfo{number}{7} (\bibinfo{year}{2010}), \bibinfo{pages}{2970--2989}.
\newblock


\bibitem[Avner and Mannor(2014)]%
        {Avner2014}
\bibfield{author}{\bibinfo{person}{Orly Avner} {and} \bibinfo{person}{Shie Mannor}.} \bibinfo{year}{2014}\natexlab{}.
\newblock \showarticletitle{Concurrent bandits and cognitive radio networks}. In \bibinfo{booktitle}{\emph{Joint European Conference on Machine Learning and Knowledge Discovery in Databases}}. \bibinfo{pages}{66--81}.
\newblock


\bibitem[Avner and Mannor(2016)]%
        {Avner2016}
\bibfield{author}{\bibinfo{person}{Orly Avner} {and} \bibinfo{person}{Shie Mannor}.} \bibinfo{year}{2016}\natexlab{}.
\newblock \showarticletitle{Multi-user lax communications: a Multi-Armed Bandit approach}. In \bibinfo{booktitle}{\emph{INFOCOM 2016-The 35th Annual IEEE International Conference on Computer Communications, IEEE}}. \bibinfo{pages}{1--9}.
\newblock


\bibitem[Bar-On and Mansour(2019)]%
        {bar2019individual}
\bibfield{author}{\bibinfo{person}{Yogev Bar-On} {and} \bibinfo{person}{Yishay Mansour}.} \bibinfo{year}{2019}\natexlab{}.
\newblock \showarticletitle{Individual Regret in Cooperative Nonstochastic Multi-Armed Bandits}. In \bibinfo{booktitle}{\emph{Adv Neural Inf Process Syst.}}, \bibfield{editor}{\bibinfo{person}{H.~Wallach}, \bibinfo{person}{H.~Larochelle}, \bibinfo{person}{A.~Beygelzimer}, \bibinfo{person}{F.~d\textquotesingle Alch\'{e}-Buc}, \bibinfo{person}{E.~Fox}, {and} \bibinfo{person}{R.~Garnett}} (Eds.), Vol.~\bibinfo{volume}{32}. \bibinfo{publisher}{Curran Associates, Inc.}, \bibinfo{pages}{3116--3126}.
\newblock


\bibitem[Bertsekas(1979)]%
        {bertsekas1979distributed}
\bibfield{author}{\bibinfo{person}{Dimitri~P Bertsekas}.} \bibinfo{year}{1979}\natexlab{}.
\newblock \showarticletitle{A distributed algorithm for the assignment problem}.
\newblock \bibinfo{journal}{\emph{Lab. for Information and Decision Systems Working Paper, MIT}} (\bibinfo{year}{1979}).
\newblock


\bibitem[Bertsekas and Castanon(1992)]%
        {bertsekas1992forward}
\bibfield{author}{\bibinfo{person}{Dimitri~P Bertsekas} {and} \bibinfo{person}{David~A Castanon}.} \bibinfo{year}{1992}\natexlab{}.
\newblock \showarticletitle{A forward/reverse auction algorithm for asymmetric assignment problems}.
\newblock \bibinfo{journal}{\emph{Computational Optimization and Applications}}  \bibinfo{volume}{1} (\bibinfo{year}{1992}), \bibinfo{pages}{277--297}.
\newblock


\bibitem[Besson and Kaufmann(2018)]%
        {Besson2018}
\bibfield{author}{\bibinfo{person}{Lilian Besson} {and} \bibinfo{person}{Emilie Kaufmann}.} \bibinfo{year}{2018}\natexlab{}.
\newblock \showarticletitle{Multi-Player Bandits Revisited}. In \bibinfo{booktitle}{\emph{Algorithmic Learning Theory}}. \bibinfo{pages}{56--92}.
\newblock


\bibitem[Bistritz et~al\mbox{.}(2020)]%
        {bistritz2020my}
\bibfield{author}{\bibinfo{person}{Ilai Bistritz}, \bibinfo{person}{Tavor Baharav}, \bibinfo{person}{Amir Leshem}, {and} \bibinfo{person}{Nicholas Bambos}.} \bibinfo{year}{2020}\natexlab{}.
\newblock \showarticletitle{My Fair Bandit: Distributed Learning of Max-Min Fairness with Multi-player Bandits}. In \bibinfo{booktitle}{\emph{Proceedings of the 37th ICML}} \emph{(\bibinfo{series}{Proceedings of Machine Learning Research}, Vol.~\bibinfo{volume}{119})}, \bibfield{editor}{\bibinfo{person}{Hal~Daumé III} {and} \bibinfo{person}{Aarti Singh}} (Eds.). \bibinfo{publisher}{PMLR}, \bibinfo{pages}{930--940}.
\newblock


\bibitem[Bistritz et~al\mbox{.}(2021)]%
        {bistritz2021one}
\bibfield{author}{\bibinfo{person}{Ilai Bistritz}, \bibinfo{person}{Tavor~Z Baharav}, \bibinfo{person}{Amir Leshem}, {and} \bibinfo{person}{Nicholas Bambos}.} \bibinfo{year}{2021}\natexlab{}.
\newblock \showarticletitle{One for All and All for One: Distributed Learning of Fair Allocations With Multi-Player Bandits}.
\newblock \bibinfo{journal}{\emph{IEEE J. Sel. Areas Inf. Theory}} \bibinfo{volume}{2}, \bibinfo{number}{2} (\bibinfo{year}{2021}), \bibinfo{pages}{584--598}.
\newblock


\bibitem[Bistritz and Bambos(2020)]%
        {bistritz2020cooperative}
\bibfield{author}{\bibinfo{person}{Ilai Bistritz} {and} \bibinfo{person}{Nicholas Bambos}.} \bibinfo{year}{2020}\natexlab{}.
\newblock \showarticletitle{Cooperative Multi-player Bandit Optimization}.
\newblock \bibinfo{journal}{\emph{Advances in Neural Information Processing Systems}}  \bibinfo{volume}{33} (\bibinfo{year}{2020}).
\newblock


\bibitem[Bistritz and Leshem(2018)]%
        {bistritz2018distributed}
\bibfield{author}{\bibinfo{person}{Ilai Bistritz} {and} \bibinfo{person}{Amir Leshem}.} \bibinfo{year}{2018}\natexlab{}.
\newblock \showarticletitle{Distributed multi-player bandits-a game of thrones approach}. In \bibinfo{booktitle}{\emph{Adv Neural Inf Process Syst.}} \bibinfo{pages}{7222--7232}.
\newblock


\bibitem[Bistritz and Leshem(2021)]%
        {bistritz2021game}
\bibfield{author}{\bibinfo{person}{Ilai Bistritz} {and} \bibinfo{person}{Amir Leshem}.} \bibinfo{year}{2021}\natexlab{}.
\newblock \showarticletitle{Game of thrones: Fully distributed learning for multiplayer bandits}.
\newblock \bibinfo{journal}{\emph{Mathematics of Operations Research}} \bibinfo{volume}{46}, \bibinfo{number}{1} (\bibinfo{year}{2021}), \bibinfo{pages}{159--178}.
\newblock


\bibitem[Boursier and Perchet(2019)]%
        {boursier2019sic}
\bibfield{author}{\bibinfo{person}{Etienne Boursier} {and} \bibinfo{person}{Vianney Perchet}.} \bibinfo{year}{2019}\natexlab{}.
\newblock \showarticletitle{SIC-MMAB: Synchronisation involves communication in multiplayer multi-armed bandits}. In \bibinfo{booktitle}{\emph{Adv Neural Inf Process Syst.}} \bibinfo{pages}{12048--12057}.
\newblock


\bibitem[Boyarski et~al\mbox{.}(2023)]%
        {boyarski2023distributed}
\bibfield{author}{\bibinfo{person}{Tomer Boyarski}, \bibinfo{person}{Wenbo Wang}, {and} \bibinfo{person}{Amir Leshem}.} \bibinfo{year}{2023}\natexlab{}.
\newblock \showarticletitle{Distributed learning for optimal spectrum access in dense device-to-device ad-hoc networks}.
\newblock \bibinfo{journal}{\emph{IEEE Transactions on Signal Processing}}  \bibinfo{volume}{71} (\bibinfo{year}{2023}), \bibinfo{pages}{3149--3163}.
\newblock


\bibitem[Bubeck et~al\mbox{.}(2012)]%
        {bubeck2012regret}
\bibfield{author}{\bibinfo{person}{S{\'e}bastien Bubeck}, \bibinfo{person}{Nicolo Cesa-Bianchi}, {et~al\mbox{.}}} \bibinfo{year}{2012}\natexlab{}.
\newblock \showarticletitle{Regret analysis of stochastic and nonstochastic multi-armed bandit problems}.
\newblock \bibinfo{journal}{\emph{Foundations and Trends{\textregistered} in Machine Learning}} \bibinfo{volume}{5}, \bibinfo{number}{1} (\bibinfo{year}{2012}), \bibinfo{pages}{1--122}.
\newblock


\bibitem[Bubeck et~al\mbox{.}(2019)]%
        {bubeck2019non}
\bibfield{author}{\bibinfo{person}{S{\'e}bastien Bubeck}, \bibinfo{person}{Yuanzhi Li}, \bibinfo{person}{Yuval Peres}, {and} \bibinfo{person}{Mark Sellke}.} \bibinfo{year}{2019}\natexlab{}.
\newblock \showarticletitle{Non-stochastic multi-player multi-armed bandits: Optimal rate with collision information, sublinear without}.
\newblock \bibinfo{journal}{\emph{arXiv preprint arXiv:1904.12233}} (\bibinfo{year}{2019}).
\newblock


\bibitem[Cohen et~al\mbox{.}(2017)]%
        {Cohen2017}
\bibfield{author}{\bibinfo{person}{Johanne Cohen}, \bibinfo{person}{Am{\'e}lie H{\'e}liou}, {and} \bibinfo{person}{Panayotis Mertikopoulos}.} \bibinfo{year}{2017}\natexlab{}.
\newblock \showarticletitle{Learning with bandit feedback in potential games}. In \bibinfo{booktitle}{\emph{Proceedings of the 31th International Conference on Neural Information Processing Systems}}.
\newblock


\bibitem[Darak and Hanawal(2019)]%
        {darak2019multi}
\bibfield{author}{\bibinfo{person}{Sumit~J Darak} {and} \bibinfo{person}{Manjesh~K Hanawal}.} \bibinfo{year}{2019}\natexlab{}.
\newblock \showarticletitle{Multi-Player Multi-Armed Bandits for Stable Allocation in Heterogeneous Ad-Hoc Networks}.
\newblock \bibinfo{journal}{\emph{IEEE Journal on Selected Areas in Communications}} \bibinfo{volume}{37}, \bibinfo{number}{10} (\bibinfo{year}{2019}), \bibinfo{pages}{2350--2363}.
\newblock


\bibitem[Evirgen and Kose(2017)]%
        {Evirgen2017}
\bibfield{author}{\bibinfo{person}{Noyan Evirgen} {and} \bibinfo{person}{Alper Kose}.} \bibinfo{year}{2017}\natexlab{}.
\newblock \showarticletitle{The Effect of Communication on Noncooperative Multiplayer Multi-Armed Bandit Problems}. In \bibinfo{booktitle}{\emph{arXiv preprint arXiv:1711.01628, 2017}}.
\newblock


\bibitem[Goldberg and Kennedy(1995)]%
        {goldberg1995efficient}
\bibfield{author}{\bibinfo{person}{Andrew~V Goldberg} {and} \bibinfo{person}{Robert Kennedy}.} \bibinfo{year}{1995}\natexlab{}.
\newblock \showarticletitle{An efficient cost scaling algorithm for the assignment problem}.
\newblock \bibinfo{journal}{\emph{Mathematical Programming}} \bibinfo{volume}{71}, \bibinfo{number}{2} (\bibinfo{year}{1995}), \bibinfo{pages}{153--177}.
\newblock


\bibitem[Hanawal and Darak(2018)]%
        {hanawal2018multi}
\bibfield{author}{\bibinfo{person}{Manjesh~K Hanawal} {and} \bibinfo{person}{Sumit~J Darak}.} \bibinfo{year}{2018}\natexlab{}.
\newblock \showarticletitle{Multi-player bandits: A trekking approach}.
\newblock \bibinfo{journal}{\emph{arXiv preprint arXiv:1809.06040}} (\bibinfo{year}{2018}).
\newblock


\bibitem[Jabbari et~al\mbox{.}(2017)]%
        {jabbari2017fairness}
\bibfield{author}{\bibinfo{person}{Shahin Jabbari}, \bibinfo{person}{Matthew Joseph}, \bibinfo{person}{Michael Kearns}, \bibinfo{person}{Jamie Morgenstern}, {and} \bibinfo{person}{Aaron Roth}.} \bibinfo{year}{2017}\natexlab{}.
\newblock \showarticletitle{Fairness in reinforcement learning}. In \bibinfo{booktitle}{\emph{Proceedings of the 34th International Conference on Machine Learning-Volume 70}}. JMLR. org, \bibinfo{pages}{1617--1626}.
\newblock


\bibitem[Joseph et~al\mbox{.}(2016)]%
        {joseph2016fairness}
\bibfield{author}{\bibinfo{person}{Matthew Joseph}, \bibinfo{person}{Michael Kearns}, \bibinfo{person}{Jamie~H Morgenstern}, {and} \bibinfo{person}{Aaron Roth}.} \bibinfo{year}{2016}\natexlab{}.
\newblock \showarticletitle{Fairness in learning: Classic and contextual bandits}. In \bibinfo{booktitle}{\emph{Advances in Neural Information Processing Systems}}. \bibinfo{pages}{325--333}.
\newblock


\bibitem[Kalai and Smorodinsky(1975)]%
        {kalai1975other}
\bibfield{author}{\bibinfo{person}{Ehud Kalai} {and} \bibinfo{person}{Meir Smorodinsky}.} \bibinfo{year}{1975}\natexlab{}.
\newblock \showarticletitle{Other solutions to Nash's bargaining problem}.
\newblock \bibinfo{journal}{\emph{Econometrica: Journal of the Econometric Society}} (\bibinfo{year}{1975}), \bibinfo{pages}{513--518}.
\newblock


\bibitem[Kalathil et~al\mbox{.}(2014)]%
        {Kalathil2014}
\bibfield{author}{\bibinfo{person}{Dileep Kalathil}, \bibinfo{person}{Naumaan Nayyar}, {and} \bibinfo{person}{Rahul Jain}.} \bibinfo{year}{2014}\natexlab{}.
\newblock \showarticletitle{Decentralized learning for multiplayer multiarmed bandits}.
\newblock \bibinfo{journal}{\emph{IEEE Transactions on Information Theory}} \bibinfo{volume}{60}, \bibinfo{number}{4} (\bibinfo{year}{2014}), \bibinfo{pages}{2331--2345}.
\newblock


\bibitem[Katz-Samuels and Jamieson(2020)]%
        {katz2020true}
\bibfield{author}{\bibinfo{person}{Julian Katz-Samuels} {and} \bibinfo{person}{Kevin Jamieson}.} \bibinfo{year}{2020}\natexlab{}.
\newblock \showarticletitle{The true sample complexity of identifying good arms}. In \bibinfo{booktitle}{\emph{International Conference on Artificial Intelligence and Statistics}}. \bibinfo{pages}{1781--1791}.
\newblock


\bibitem[Kubiak(2008)]%
        {kubiak2008proportional}
\bibfield{author}{\bibinfo{person}{Wieslaw Kubiak}.} \bibinfo{year}{2008}\natexlab{}.
\newblock \bibinfo{booktitle}{\emph{Proportional optimization and fairness}}. Vol.~\bibinfo{volume}{127}.
\newblock \bibinfo{publisher}{Springer Science \& Business Media}.
\newblock


\bibitem[Lai et~al\mbox{.}(2008)]%
        {Lai2008}
\bibfield{author}{\bibinfo{person}{Lifeng Lai}, \bibinfo{person}{Hai Jiang}, {and} \bibinfo{person}{H~Vincent Poor}.} \bibinfo{year}{2008}\natexlab{}.
\newblock \showarticletitle{Medium access in cognitive radio networks: A competitive multi-armed bandit framework}. In \bibinfo{booktitle}{\emph{Signals, Systems and Computers, 2008 42nd Asilomar Conference on}}. \bibinfo{pages}{98--102}.
\newblock


\bibitem[Lai and Robbins(1984)]%
        {lai1984asymptotically}
\bibfield{author}{\bibinfo{person}{Tze~Leung Lai} {and} \bibinfo{person}{Herbert Robbins}.} \bibinfo{year}{1984}\natexlab{}.
\newblock \showarticletitle{Asymptotically optimal allocation of treatments in sequential experiments}.
\newblock \bibinfo{journal}{\emph{Design of Experiments: Ranking and Selection}} (\bibinfo{year}{1984}), \bibinfo{pages}{127--142}.
\newblock


\bibitem[Lai and Robbins(1985)]%
        {lai1985asymptotically}
\bibfield{author}{\bibinfo{person}{Tze~Leung Lai} {and} \bibinfo{person}{Herbert Robbins}.} \bibinfo{year}{1985}\natexlab{}.
\newblock \showarticletitle{Asymptotically efficient adaptive allocation rules}.
\newblock \bibinfo{journal}{\emph{Advances in applied mathematics}} \bibinfo{volume}{6}, \bibinfo{number}{1} (\bibinfo{year}{1985}), \bibinfo{pages}{4--22}.
\newblock


\bibitem[Liu et~al\mbox{.}(2013)]%
        {Liu2013}
\bibfield{author}{\bibinfo{person}{Haoyang Liu}, \bibinfo{person}{Keqin Liu}, {and} \bibinfo{person}{Qing Zhao}.} \bibinfo{year}{2013}\natexlab{}.
\newblock \showarticletitle{Learning in a changing world: Restless multiarmed bandit with unknown dynamics}.
\newblock \bibinfo{journal}{\emph{IEEE Transactions on Information Theory}} \bibinfo{volume}{59}, \bibinfo{number}{3} (\bibinfo{year}{2013}), \bibinfo{pages}{1902--1916}.
\newblock


\bibitem[Liu and Zhao(2010)]%
        {liu2010distributed}
\bibfield{author}{\bibinfo{person}{Keqin Liu} {and} \bibinfo{person}{Qing Zhao}.} \bibinfo{year}{2010}\natexlab{}.
\newblock \showarticletitle{Distributed learning in multi-armed bandit with multiple players}.
\newblock \bibinfo{journal}{\emph{{IEEE} Trans. Signal Process.}} \bibinfo{volume}{58}, \bibinfo{number}{11} (\bibinfo{year}{2010}), \bibinfo{pages}{5667--5681}.
\newblock


\bibitem[Liu et~al\mbox{.}(2019)]%
        {liu2019competing}
\bibfield{author}{\bibinfo{person}{Lydia~T Liu}, \bibinfo{person}{Horia Mania}, {and} \bibinfo{person}{Michael~I Jordan}.} \bibinfo{year}{2019}\natexlab{}.
\newblock \showarticletitle{Competing bandits in matching markets}.
\newblock \bibinfo{journal}{\emph{arXiv preprint arXiv:1906.05363}} (\bibinfo{year}{2019}).
\newblock


\bibitem[Mehrabian et~al\mbox{.}(2020)]%
        {boursier2019practical}
\bibfield{author}{\bibinfo{person}{Abbas Mehrabian}, \bibinfo{person}{Etienne Boursier}, \bibinfo{person}{Emilie Kaufmann}, {and} \bibinfo{person}{Vianney Perchet}.} \bibinfo{year}{2020}\natexlab{}.
\newblock \showarticletitle{A practical algorithm for multiplayer bandits when arm means vary among players}. In \bibinfo{booktitle}{\emph{23rd AISTATS}}. PMLR, \bibinfo{address}{online}, \bibinfo{pages}{1211--1221}.
\newblock


\bibitem[Mjelde(1983)]%
        {mjelde1983properties}
\bibfield{author}{\bibinfo{person}{K{\aa}re~M Mjelde}.} \bibinfo{year}{1983}\natexlab{}.
\newblock \showarticletitle{Properties of Pareto optimal allocations of resources to activities}.
\newblock \bibinfo{journal}{\emph{modeling, identification and control}} \bibinfo{volume}{4}, \bibinfo{number}{3} (\bibinfo{year}{1983}), \bibinfo{pages}{167--173}.
\newblock


\bibitem[Mo and Walrand(2000)]%
        {mo2000fair_alpha}
\bibfield{author}{\bibinfo{person}{Jeonghoon Mo} {and} \bibinfo{person}{Jean Walrand}.} \bibinfo{year}{2000}\natexlab{}.
\newblock \showarticletitle{Fair end-to-end window-based congestion control}.
\newblock \bibinfo{journal}{\emph{IEEE/ACM Transactions on networking}} \bibinfo{number}{5} (\bibinfo{year}{2000}), \bibinfo{pages}{556--567}.
\newblock


\bibitem[Naparstek and Leshem(2014)]%
        {naparstek2013fully}
\bibfield{author}{\bibinfo{person}{Oshri Naparstek} {and} \bibinfo{person}{Amir Leshem}.} \bibinfo{year}{2014}\natexlab{}.
\newblock \showarticletitle{Fully distributed optimal channel assignment for open spectrum access}.
\newblock \bibinfo{journal}{\emph{{IEEE} Trans. Signal Process.}} \bibinfo{volume}{62}, \bibinfo{number}{2} (\bibinfo{year}{2014}), \bibinfo{pages}{283--294}.
\newblock


\bibitem[Naparstek and Leshem(2016)]%
        {naparstek2016expected}
\bibfield{author}{\bibinfo{person}{Oshri Naparstek} {and} \bibinfo{person}{Amir Leshem}.} \bibinfo{year}{2016}\natexlab{}.
\newblock \showarticletitle{Expected time complexity of the auction algorithm and the push relabel algorithm for maximum bipartite matching on random graphs}.
\newblock \bibinfo{journal}{\emph{Random Structures \& Algorithms}} \bibinfo{volume}{48}, \bibinfo{number}{2} (\bibinfo{year}{2016}), \bibinfo{pages}{384--395}.
\newblock


\bibitem[Nash~Jr(1950)]%
        {nash1950bargaining}
\bibfield{author}{\bibinfo{person}{John~F Nash~Jr}.} \bibinfo{year}{1950}\natexlab{}.
\newblock \showarticletitle{The bargaining problem}.
\newblock \bibinfo{journal}{\emph{Econometrica: Journal of the econometric society}} (\bibinfo{year}{1950}), \bibinfo{pages}{155--162}.
\newblock


\bibitem[Nayyar et~al\mbox{.}(2016)]%
        {nayyar2016regret}
\bibfield{author}{\bibinfo{person}{Naumaan Nayyar}, \bibinfo{person}{Dileep Kalathil}, {and} \bibinfo{person}{Rahul Jain}.} \bibinfo{year}{2016}\natexlab{}.
\newblock \showarticletitle{On regret-optimal learning in decentralized multiplayer multiarmed bandits}.
\newblock \bibinfo{journal}{\emph{{IEEE} Trans. Control Netw. Syst.}} \bibinfo{volume}{5}, \bibinfo{number}{1} (\bibinfo{year}{2016}), \bibinfo{pages}{597--606}.
\newblock


\bibitem[Radunovic and Le~Boudec(2007)]%
        {radunovic2007unified}
\bibfield{author}{\bibinfo{person}{Bozidar Radunovic} {and} \bibinfo{person}{Jean-Yves Le~Boudec}.} \bibinfo{year}{2007}\natexlab{}.
\newblock \showarticletitle{A unified framework for max-min and min-max fairness with applications}.
\newblock \bibinfo{journal}{\emph{IEEE/ACM Transactions on networking}} \bibinfo{volume}{15}, \bibinfo{number}{5} (\bibinfo{year}{2007}), \bibinfo{pages}{1073--1083}.
\newblock


\bibitem[Ramanathan et~al\mbox{.}(2007)]%
        {Ramanathan2007RandomizedLeaderElection}
\bibfield{author}{\bibinfo{person}{Murali~Krishna Ramanathan}, \bibinfo{person}{Ronaldo~A. Ferreira}, \bibinfo{person}{Suresh Jagannathan}, \bibinfo{person}{Ananth Grama}, {and} \bibinfo{person}{Wojciech Szpankowski}.} \bibinfo{year}{2007}\natexlab{}.
\newblock \showarticletitle{Randomized Leader Election}.
\newblock \bibinfo{journal}{\emph{Distributed Computing}}  \bibinfo{volume}{19} (\bibinfo{year}{2007}), \bibinfo{pages}{403--418}.
\newblock
\href{https://doi.org/10.1007/s00446-007-0022-4}{doi:\nolinkurl{10.1007/s00446-007-0022-4}}


\bibitem[Rom and Sidi(2012)]%
        {rom2012multiple}
\bibfield{author}{\bibinfo{person}{Raphael Rom} {and} \bibinfo{person}{Moshe Sidi}.} \bibinfo{year}{2012}\natexlab{}.
\newblock \bibinfo{booktitle}{\emph{Multiple access protocols: Performance and analysis}}.
\newblock \bibinfo{publisher}{Springer Science \& Business Media}.
\newblock


\bibitem[Rosenski et~al\mbox{.}(2016)]%
        {Rosenski2016}
\bibfield{author}{\bibinfo{person}{Jonathan Rosenski}, \bibinfo{person}{Ohad Shamir}, {and} \bibinfo{person}{Liran Szlak}.} \bibinfo{year}{2016}\natexlab{}.
\newblock \showarticletitle{Multi-player bandits--a musical chairs approach}. In \bibinfo{booktitle}{\emph{International Conference on Machine Learning}}. \bibinfo{pages}{155--163}.
\newblock


\bibitem[Tibrewal et~al\mbox{.}(2019)]%
        {tibrewal2019distributed}
\bibfield{author}{\bibinfo{person}{Harshvardhan Tibrewal}, \bibinfo{person}{Sravan Patchala}, \bibinfo{person}{Manjesh~K Hanawal}, {and} \bibinfo{person}{Sumit~J Darak}.} \bibinfo{year}{2019}\natexlab{}.
\newblock \showarticletitle{Distributed Learning and Optimal Assignment in Multiplayer Heterogeneous Networks}. In \bibinfo{booktitle}{\emph{IEEE INFOCOM 2019-IEEE Conference on Computer Communications}}. IEEE, \bibinfo{pages}{1693--1701}.
\newblock


\bibitem[Vakili et~al\mbox{.}(2013)]%
        {Vakili2013}
\bibfield{author}{\bibinfo{person}{Sattar Vakili}, \bibinfo{person}{Keqin Liu}, {and} \bibinfo{person}{Qing Zhao}.} \bibinfo{year}{2013}\natexlab{}.
\newblock \showarticletitle{Deterministic sequencing of exploration and exploitation for multi-armed bandit problems}.
\newblock \bibinfo{journal}{\emph{IEEE Journal of Selected Topics in Signal Processing}} \bibinfo{volume}{7}, \bibinfo{number}{5} (\bibinfo{year}{2013}), \bibinfo{pages}{759--767}.
\newblock


\bibitem[Xu et~al\mbox{.}(2015)]%
        {xu2015distributed}
\bibfield{author}{\bibinfo{person}{Jie Xu}, \bibinfo{person}{Cem Tekin}, \bibinfo{person}{Simpson Zhang}, {and} \bibinfo{person}{Mihaela Van Der~Schaar}.} \bibinfo{year}{2015}\natexlab{}.
\newblock \showarticletitle{Distributed multi-agent online learning based on global feedback}.
\newblock \bibinfo{journal}{\emph{IEEE Transactions on Signal Processing}} \bibinfo{volume}{63}, \bibinfo{number}{9} (\bibinfo{year}{2015}), \bibinfo{pages}{2225--2238}.
\newblock


\bibitem[Zafaruddin et~al\mbox{.}(2019)]%
        {zafaruddin2019distributed}
\bibfield{author}{\bibinfo{person}{SM Zafaruddin}, \bibinfo{person}{Ilai Bistritz}, \bibinfo{person}{Amir Leshem}, {and} \bibinfo{person}{Dusit Niyato}.} \bibinfo{year}{2019}\natexlab{}.
\newblock \showarticletitle{Distributed learning for channel allocation over a shared spectrum}.
\newblock \bibinfo{journal}{\emph{{IEEE} J. Sel. Areas Commun.}} \bibinfo{volume}{37}, \bibinfo{number}{10} (\bibinfo{year}{2019}), \bibinfo{pages}{2337--2349}.
\newblock


\bibitem[Zehavi et~al\mbox{.}(2013)]%
        {zehavi2013weighted}
\bibfield{author}{\bibinfo{person}{Ephraim Zehavi}, \bibinfo{person}{Amir Leshem}, \bibinfo{person}{Ronny Levanda}, {and} \bibinfo{person}{Zhu Han}.} \bibinfo{year}{2013}\natexlab{}.
\newblock \showarticletitle{Weighted max-min resource allocation for frequency selective channels}.
\newblock \bibinfo{journal}{\emph{IEEE transactions on signal processing}} \bibinfo{volume}{61}, \bibinfo{number}{15} (\bibinfo{year}{2013}), \bibinfo{pages}{3723--3732}.
\newblock


\bibitem[Zhang et~al\mbox{.}(2019)]%
        {zhang2019group}
\bibfield{author}{\bibinfo{person}{Xueru Zhang}, \bibinfo{person}{Mohammadmahdi Khaliligarekani}, \bibinfo{person}{Cem Tekin}, {et~al\mbox{.}}} \bibinfo{year}{2019}\natexlab{}.
\newblock \showarticletitle{Group retention when using machine learning in sequential decision making: the interplay between user dynamics and fairness}.
\newblock \bibinfo{journal}{\emph{Advances in neural information processing systems}}  \bibinfo{volume}{32} (\bibinfo{year}{2019}).
\newblock


\bibitem[Zhao and Tong(2005)]%
        {zhao2005opportunistic}
\bibfield{author}{\bibinfo{person}{Qing Zhao} {and} \bibinfo{person}{Lang Tong}.} \bibinfo{year}{2005}\natexlab{}.
\newblock \showarticletitle{Opportunistic carrier sensing for energy-efficient information retrieval in sensor networks}.
\newblock \bibinfo{journal}{\emph{EURASIP J. Wirel. Commun. Netw.}} \bibinfo{volume}{2005}, \bibinfo{number}{2} (\bibinfo{year}{2005}), \bibinfo{pages}{1--11}.
\newblock


\end{thebibliography}

\appendix
\section{Pseudo-code for the proposed algorithm}
\label{app:code}
We present the pseudo-code of the proposed algorithm.
The general structure is agent ordering followed by epochs each consisting of three phases. The length of each phase is determined as described in the paper. 
Algorithm~\ref{alg:FairBandits} in the main text gives the high-level structure; the following algorithms describe the individual phases. 
\subsection{Agent Ordering}
\label{app:agent_ordering}
The goal of the ordering phase is to assign each agent a distinct temporary
rank $i_n\in\{1,\ldots,N\}$, which is identified with one of the first $N$
arms. The phase proceeds in even--odd pairs of slots. At an even slot,
each unassigned agent randomly selects one of the arms that it still regards
as available. Assigned agents access only their assigned arms. If an
unassigned agent selects an arm and observes no collision, it declares
itself assigned to that arm.

If an unassigned agent observes a collision at an even slot, the collision
may have been caused either by an already assigned agent occupying that arm,
or by another unassigned agent selecting the same arm. To distinguish these
two cases, the unassigned agent repeats the same arm in the following odd
slot, while assigned agents remain silent. If no collision occurs in the odd
slot, the agent concludes that the previous collision was with an assigned
agent and marks that arm as unavailable. If a collision occurs again, then
the collision was with another unassigned agent, and the arm is not marked
unavailable.

At the end of each ordering block, the agents run a termination check. Each
unassigned agent scans all arms according to the common order. Assigned
agents access their assigned arms in their corresponding slots. Therefore,
if any unassigned agent remains, every assigned agent experiences a collision
during this scan and learns that the ordering phase has not ended. If no
collision occurs in the termination check, all agents know that every agent
has been assigned a distinct rank, and the ordering phase terminates.

The duration of one ordering block can be justified by a coupon-collector
argument. The bottleneck occurs when only one agent remains unassigned.
At this point there is exactly one unassigned arm. In each even--odd pair,
the remaining unassigned agent selects one of the arms that it still regards
as available. If it selects the unique unassigned arm, it experiences no
collision and becomes assigned. If it selects an already assigned arm, then
the odd-slot repetition reveals that the collision was with an assigned
agent, and the arm is marked unavailable. Thus the agent either becomes
assigned or removes one unavailable arm from its candidate set.
Even ignoring the removal of unavailable arms, the probability that the last unassigned agent does not select the unique free arm in
\(L\) independent trials is at most
\[
\left(1-\frac{1}{N}\right)^L
\le \exp\!\left(-\frac{L}{N}\right).
\]
Taking \(L=4N\log_2 N\), this probability is bounded by
\begin{equation}
\exp(-4\log_2 N)
=
N^{-4/\ln 2}.
\end{equation}
Hence, a block of \(4N\log_2 N\) even--odd trials assigns the last remaining
agent with probability at least \(1-N^{-4/\ln 2}\). Since the last agent
case is the slowest, this gives a high-probability justification for
the ordering-block length. The termination scan following the block ensures that if any agent remains unassigned, all assigned agents observe a collision
and the ordering block is repeated.
\newpage
\begin{algorithm}[H]
\caption{\texttt{Agent ordering(n)}}
\label{alg:ordering}
\begin{algorithmic}[1]
\STATE \textbf{Initialization}
\STATE endflag $\gets 0$.
\STATE Each agent $n$ sets:
\STATE $s_n \gets$ unassigned.
\STATE Set arm $m$ available $\gets 1$, $m=1,\ldots,N$.
\STATE Note that we only need the first $N$ arms for ordering.
\WHILE {endflag=0}
     \FOR {$t=0,\ldots 4N\log_2(N)-1$}
        \IF {$t$ is even}
            \IF {$s_n=$ unassigned}
                \STATE Randomly select available arm $m_n(t)$. 
                \STATE Sample arm $m_n(t)$.
                \IF {$r_{n,m_n(t)}\neq 0$}
                    \STATE $s_n \gets$ assigned.
                    \STATE $i_n=m_n(t)$.
                \ENDIF 
            \ELSE
                \STATE sample arm $i_n$.
            \ENDIF
        \ELSE
            \IF {$s_n=$ unassigned}
                \STATE Sample again arm $m(t-1)$.
                \IF {$r_{n,m(t-1)} \neq 0$}
                    \STATE arm $m(t-1)$ $\gets$ unavailable.
                \ENDIF
            \ENDIF
        \ENDIF 
    \ENDFOR
    \FOR {$t=1:N$}
        \IF {$s_n=$unassigned}
            \STATE Sample arm $t$.
        \ELSE
            \IF {$t=i_n$}
                \STATE Sample arm $i_n$.
                \IF{$r_{n,i_n}\neq 0$}
                    \STATE endflag $\gets 1$.
                \ENDIF    
            \ENDIF
        \ENDIF 
    \ENDFOR
\ENDWHILE
\STATE Output for agent $n$ is a number $1\le i_n \le N$.
\end{algorithmic}
\end{algorithm}
\subsection{Exploration}
The exploration procedure is presented in Algorithm~\ref{alg:exploration}. In epoch $k$, each agent collects
$\Delta \ell_k=L\bigl(g(k)-g(k-1)\bigr),
    g(k)=kf(k),
$
new samples from each arm. The ordered agents use a deterministic round-robin schedule, so no collisions occur during exploration.

\begin{algorithm}[H]
\caption{\texttt{Exploration$(k,n)$}}
\label{alg:exploration}
\begin{algorithmic}[1]
\STATE \textbf{Input:} epoch $k$, rank $i_n$, schedule $f$, parameter $L$.
\STATE $g(k)\gets kf(k)$ and $g(k-1)\gets (k-1)f(k-1)$, with $g(0)=0$.
\STATE $\Delta\ell_k\gets L\bigl(g(k)-g(k-1)\bigr)$.
\FOR{$q=1$ to $\Delta\ell_k$}
    \FOR{$j=1$ to $M$}
        \STATE $m\gets ((i_n+j+q-2)\bmod M)+1$.
        \STATE Agent $n$ samples arm $m$ and observes $r_{n,m}$.
        \STATE $V_{n,m}\gets V_{n,m}+1$.
        \STATE $\Rh_{n,m}\gets \Rh_{n,m}+\frac{1}{V_{n,m}}\bigl(r_{n,m}-\Rh_{n,m}\bigr)$.
    \ENDFOR
\ENDFOR
\STATE \textbf{Output:} updated $\Rh_{n,m}$, $V_{n,m}$ for all $m$.
\end{algorithmic}
\end{algorithm}
\newpage
\subsection{Matching}
The matching phase implements the endpoint-revalidated warm-started bisection described in Section~\ref{sec:matching_revised}. Each feasibility test invokes the distributed auction routine in Algorithm~\ref{alg:dist_auction}.
\begin{algorithm}[H]
\caption{Endpoint revalidation and local bracket repair$(k,n)$}
\label{alg:matching_endpoint_repair}
\begin{algorithmic}[1]
\STATE Input: empirical rewards $\Rh_{n,m}(k)$ and previous bracket
$[\tau_{\min}(k-1),\tau_{\max}(k-1)]$.
\IF{$k=1$ or no valid bracket is stored}
    \STATE $\tau_{\min}\gets 0$, $\tau\gets 1$, $\mathrm{Infeasible}\gets 0$.
    \WHILE{$\mathrm{Infeasible}=0$}
        \STATE Build $A_{n,m}(\tau)=\mathbf 1\{\Rh_{n,m}(k)\ge \tau\}$.
        \STATE $\pi_n\gets \mathrm{DistributedAuction}(A_n(\tau))$.
        \IF{$\pi_n=0$}
            \STATE $\mathrm{Infeasible}\gets 1$; $\tau_{\max}\gets \tau$.
        \ELSE
            \STATE $\tau_{\min}\gets \tau$; $\mathrm{Allocation}(n)\gets \pi_n$; $\tau\gets 2\tau$.
        \ENDIF
    \ENDWHILE
\ELSE
    \STATE Test feasibility at $\tau_{\min}(k-1)$ using $\Rh(k)$.
    \STATE Test feasibility at $\tau_{\max}(k-1)$ using $\Rh(k)$.
    \STATE $\eta\gets \max\{\tau_{\max}(k-1)-\tau_{\min}(k-1),2^{-f(k-1)}\}$.
    \IF{$\tau_{\min}(k-1)$ feasible and $\tau_{\max}(k-1)$ infeasible}
        \STATE $\tau_{\min}\gets\tau_{\min}(k-1)$;
        $\tau_{\max}\gets\tau_{\max}(k-1)$.
    \ELSIF{$\tau_{\max}(k-1)$ feasible}
        \STATE $\tau_{\min}\gets\tau_{\max}(k-1)$; $q\gets0$.
        \REPEAT
            \STATE $\tau\gets\tau_{\min}+2^q\eta$.
            \STATE Test feasibility at $\tau$ using $\Rh(k)$.
            \STATE $q\gets q+1$.
        \UNTIL{$\tau$ is infeasible}
        \STATE $\tau_{\max}\gets\tau$.
    \ELSIF{$\tau_{\min}(k-1)$ infeasible}
        \STATE $\tau_{\max}\gets\tau_{\min}(k-1)$; $q\gets0$.
        \REPEAT
            \STATE $\tau\gets\max\{0,\tau_{\max}-2^q\eta\}$.
            \STATE Test feasibility at $\tau$ using $\Rh(k)$.
            \STATE $q\gets q+1$.
        \UNTIL{$\tau$ is feasible}
        \STATE $\tau_{\min}\gets\tau$.
    \ENDIF
\ENDIF
\STATE Output: a valid bracket $[\tau_{\min},\tau_{\max}]$ for epoch $k$.
\end{algorithmic}
\end{algorithm}
\begin{algorithm}[H]
\caption{Endpoint-revalidated distributed matching phase$(k,n)$}
\label{alg:matching_endpoint_revalidation}
\begin{algorithmic}[1]
\STATE Input: empirical rewards $\Rh_{n,m}(k)$, a valid bracket
$[\tau_{\min},\tau_{\max}]$ from Algorithm~\ref{alg:matching_endpoint_repair}, and previous allocation $\mathrm{Allocation}(n)$.
\FOR{$j=1$ to $s(k)=\lceil f(k)\rceil$}
    \STATE $\tau\gets(\tau_{\min}+\tau_{\max})/2$.
    \STATE Build $A_{n,m}(\tau)=\mathbf 1\{\Rh_{n,m}(k)\ge \tau\}$.
    \STATE $\pi_n\gets \mathrm{DistributedAuction}(A_n(\tau))$.
    \IF{$\pi_n=0$}
        \STATE $\tau_{\max}\gets\tau$.
    \ELSE
        \STATE $\tau_{\min}\gets\tau$; $\mathrm{Allocation}(n)\gets\pi_n$.
    \ENDIF
\ENDFOR
\STATE $w_k\gets 2^{-f(k)}$ and $d\gets\tau_{\max}-\tau_{\min}$.
\IF{$d<w_k$}
    \STATE $m\gets w_k-d$ and $a\gets\min\{\tau_{\min},m/2\}$.
    \STATE $\tau_{\min}\gets\tau_{\min}-a$.
    \STATE $\tau_{\max}\gets\tau_{\max}+m-a$.
\ENDIF
\STATE Store $[\tau_{\min}(k),\tau_{\max}(k)]\gets[\tau_{\min},\tau_{\max}]$.
\STATE Output: $\mathrm{Allocation}(n)$.
\end{algorithmic}
\end{algorithm}
\begin{algorithm}[H]
\caption{\texttt{Distributed auction(n)}}
\label{alg:dist_auction}
\begin{algorithmic}[1]
\STATE \textbf{Input}: 
    \STATE $A_{n,m}, m=1,\ldots,M$.
\STATE \textbf{Initialization}
\STATE Assigned$(n) \gets 0$.
\FOR {$m=1,\dots,M$}
     \IF {$A_{n,m}=1$} 
        \STATE $h_{n,m} \gets 0$
     \ELSE
        \STATE $h_{n,m} \gets \infty$
     \ENDIF
\ENDFOR
\FOR {$t=1,\ldots N^2(N-1)$}
    \IF {Assigned$(n)=1$}
        \STATE Access arm $\pi(n)$.
        \IF {$r_{n,\pi(n)}=0$}
            \STATE Assigned$(n) \gets 0$.
            \STATE $h_{n,\pi(n)} \gets h_{n,\pi(n)}+1$.
            \STATE $\pi(n)\gets 0$.
        \ENDIF
    \ELSE 
    \IF {$t \equiv i_n \pmod N$}
       \IF {$\min_m h_{n,m}<\infty$ }
       \STATE $\pi(n)\leftarrow \arg\min_m h_{n,m}$
       \STATE Access arm $\pi(n)$
     \STATE Assigned$(n)\leftarrow 1$
    \ELSE
    \STATE Assigned$(n)\leftarrow 0$
     \ENDIF  
     \ENDIF    
     \ENDIF
\ENDFOR
\FOR {$t=1,\ldots,N$}
    \IF {Assigned$(n)=0$}
        \STATE Access arm $t$ 
   \ELSE 
        \IF {$t=\pi(n)$}
            \STATE Access arm $\pi(n)$
            \IF {$r_{n,\pi(n)}=0$}
                \STATE $\pi(n) \gets 0$
            \ENDIF
        \ENDIF
    \ENDIF
\ENDFOR
\STATE  \textbf{Output:}
\STATE  $\pi(n)$. If $\pi(n)=0$, then the threshold graph has no perfect matching. 
\end{algorithmic}
\end{algorithm}
\subsection{Exploitation}
In the exploitation phase, each agent repeatedly plays the allocation returned by the most recent feasible matching test. In the analysis, the exploitation length in epoch $k$ is $c_3(k)=2^k$, although other exponential bases can be used as tuning parameters. At the end of the phase, the agents use the ordered collision schedule to synchronize and move to the next epoch.

\end{document}